\documentclass{article}

\PassOptionsToPackage{numbers,compress}{natbib}
\usepackage[main, final]{neurips_2026}

\usepackage[utf8]{inputenc}
\usepackage[T1]{fontenc}
\usepackage{hyperref}
\usepackage{url}
\usepackage{booktabs}
\usepackage{amsfonts}
\usepackage{nicefrac}
\usepackage{microtype}
\usepackage{graphicx}
\usepackage{multirow}
\usepackage{amsmath}
\usepackage{amssymb}
\usepackage{xcolor}
\usepackage{float}
\usepackage{bbm}
\usepackage{enumitem}
\usepackage{placeins}

\makeatletter
\renewcommand{\section}{%
  \@startsection{section}{1}{\z@}%
                {-1.5ex \@plus -0.5ex \@minus -0.2ex}%
                {0.8ex \@plus 0.2ex}%
                {\large\bf\raggedright}}
\renewcommand{\subsection}{%
  \@startsection{subsection}{2}{\z@}%
                {-1.2ex \@plus -0.4ex \@minus -0.2ex}%
                {0.5ex \@plus 0.1ex}%
                {\normalsize\bf\raggedright}}
\renewcommand{\subsubsection}{%
  \@startsection{subsubsection}{3}{\z@}%
                {-1.0ex \@plus -0.3ex \@minus -0.2ex}%
                {0.3ex \@plus 0.1ex}%
                {\normalsize\bf\raggedright}}
\makeatother

\title{Multimodal LLMs Outperform Pathology Foundation Models in Cross-Domain Histological Similarity}

\author{
  Yishu Zhang \quad Yun Li \quad Daiwei Zhang\thanks{Corresponding author.} \\
  University of North Carolina at Chapel Hill \\
  \texttt{yzhang4@unc.edu}, \texttt{yun\_li@med.unc.edu}, \texttt{daiwei@unc.edu}
}

\begin{document}

\maketitle

\begin{abstract}
State-of-the-art pathology foundation models, trained on millions of histology tiles, can fail to preserve tissue similarity when comparisons cross slide or institution boundaries. We show that general-purpose multimodal LLMs, without being trained as pathology foundation models, consistently outperform these specialized models in cross-domain histological similarity judgments. Using a relative similarity framework that we release as the MOSAIC (Model Similarity Assessment across Institutions and Cohorts) benchmark, we evaluate 17 models across 6 datasets and find that pathology encoders often rank same-institution, different-disease tiles as more similar than same-disease, different-institution tiles, a clinically dangerous failure mode invisible to standard within-domain evaluations. LLMs appear less susceptible to this failure, likely because they perform semantic visual comparison of morphology and tissue architecture rather than relying on shortcut features tied to acquisition context. Scaling training data does not resolve the problem for pathology encoders, implicating the learning objective rather than data coverage. Our results expose a fundamental robustness gap in current pathology foundation models and establish multimodal LLMs as a viable alternative for cross-institutional retrieval, dataset harmonization, and multi-site quality control. Code and data will be released upon acceptance.
\end{abstract}

\section{Introduction}
\label{sec:introduction}

Large-scale self-supervised pretraining has transformed computational pathology through visual foundation models~\citep{chen2024uni, zimmermann2024virchow2,filiot2024phikon, xu2024gigapath, nechaev2024hibou}. Trained on millions of histology image tiles, these models extract general-purpose representations that support a broad spectrum of downstream tasks, including tissue classification, cancer subtyping, and biomarker inference~\cite{filiot2024eva, campanella2025benchmark}. For these representations to support safety-critical applications such as cross-institutional image retrieval, few-shot adaptation, and multi-site clinical trials, the learned embedding space must cluster images by disease-relevant morphology rather than by acquisition context.

In this work, we expose a fundamental failure mode of current pathology foundation models: their embedding spaces can remain strongly entangled with technical variation, causing them to rank same-institution, different-disease tiles as more similar than same-disease, different-institution tiles. This failure is invisible to standard within-domain evaluations (linear probes, $k$-NN) yet directly undermines the retrieval and transfer tasks that motivate foundation model development. Concretely, if a pathologist queries a multi-site case database, a model with this failure mode will return color-matched tissue from the same scanner rather than morphologically equivalent disease from another laboratory, with direct consequences for diagnostic support and clinical decision-making.

Recent work has confirmed that pathology foundation model embeddings encode site-level technical signatures more strongly than biological class~\citep{komen2024batch, deJong2025, komen2025towards}, and that standard stain normalization does not fully remove these shortcuts~\citep{Howard2021, hoque2024stain, chai2026impact}. However, prior evaluations rely on embedding geometry or downstream classifiers, making them inapplicable to generative models. Standard task-specific probes (linear classifiers, $k$-NN) can further obscure the problem by achieving high accuracy through cohort-specific correlations~\citep{Dehkharghanian2023,lin2025impact}.

To expose and quantify this failure mode, we introduce a relative similarity framework, inspired by representational similarity analysis~\cite{kriegeskorte2008rsa} and human similarity judgments~\cite{roads2021enriching, muttenthaler2025aligning}, that tests whether a model ranks a biologically similar tile from a different source (slide or institution) above a biologically dissimilar tile that shares the same acquisition context. This formulation can be applied uniformly to both embedding-based models and generative LLMs. Using this framework, we demonstrate that multimodal LLMs, without being trained as pathology foundation models, consistently outperform specialized pathology encoders in cross-domain settings, serving as an existence proof that the failure is specific to current training paradigms rather than inherent to the task.

Our contributions are: (1) We expose a clinically dangerous robustness gap in state-of-the-art pathology foundation models that is invisible to standard evaluations. (2) We introduce a relative similarity evaluation framework that directly quantifies cross-domain generalization. (3) We demonstrate that multimodal LLMs serve as an existence proof that robust cross-domain similarity is achievable without domain-specific training. (4) We release MOSAIC, a benchmark suite including comparison triplets, evaluation code, and baseline outputs, spanning 6 histopathology cohorts for evaluating cross-domain generalization in computational pathology.

\section{Related Work}
\label{sec:related_work}

\textbf{Robustness Evaluation for Pathology Models.} K\"{o}men et al.~\citep{komen2024batch} demonstrated that pathology foundation model embeddings retain hospital-specific signatures that persist across principal components even after stain normalization. De Jong et al.~\citep{deJong2025} formalized this with the Robustness Index, which quantifies whether embedding neighborhoods are dominated more by biological labels or medical center labels, finding that nearly all evaluated models encode site identity more strongly than tissue type. PathoROB~\citep{komen2025towards} scaled this analysis to 20 models across 34 centers, establishing systematic benchmarking of embedding robustness. Additional work has examined scanner-specific~\citep{schirris2026scanner} and staining-specific~\citep{chai2026impact} domain shifts. All of these approaches operate on embedding geometry or downstream classifiers, requiring access to continuous vector representations. Our relative similarity framework complements these methods by reformulating robustness evaluation as a comparative judgment task, enabling direct comparison between embedding-based models and generative multimodal LLMs on the same task.

\textbf{Domain Shift Mitigation in Computational Pathology.}
Multiple strategies have been proposed to reduce the impact of technical variation on histopathology models. Stain normalization methods transform images to a canonical color space~\citep{hoque2024stain}, while stain augmentation approaches like RandStainNA~\citep{shen2022randstainna} train models to be invariant to stain variation by sampling diverse styles during training. Style transfer methods such as STRAP~\citep{yamashita2021strap} use medically irrelevant artistic images as style sources to learn domain-agnostic representations. However, these methods target color and style variation, whereas the robustness studies above show that residual site information extends beyond color to deeper structural artifacts (tissue thickness, cutting angle, scanner optics) that current mitigation strategies do not fully address~\citep{komen2024batch, tellez2019quantifying}. Our work takes a complementary perspective: rather than modifying the input to remove domain shift, we evaluate whether models can make correct similarity judgments despite it.

\textbf{Representational Similarity and Human Judgment.}
Representational similarity analysis (RSA)~\citep{kriegeskorte2008rsa} provides a framework for comparing model representations by correlating dissimilarity matrices, independent of task-specific performance. Human similarity judgments offer a principled ground truth for evaluating representational geometry: Hebart et al.~\citep{hebart2020things} collected 1.5 million odd-one-out triplet judgments to reveal the dimensional structure of object representations, and Roads and Love~\citep{roads2021enriching} enriched ImageNet with human similarity data. Muttenthaler et al.~\citep{muttenthaler2023improving} showed that aligning neural network representations with human triplet judgments improves downstream generalization, and subsequent work demonstrated that such alignment transfers across abstraction levels~\citep{muttenthaler2025aligning}. Our relative similarity framework adapts the triplet evaluation paradigm to histopathology, using tissue-class structure as the ground truth rather than human perceptual judgments, and specifically tests whether similarity is preserved across domain boundaries. We operationalize this as MOSAIC, a cross-domain triplet benchmark for tissue-level biological similarity.

\textbf{Multimodal LLMs as Visual Judges.}
The LLM-as-a-judge paradigm~\citep{zheng2023judging} has established that language models can serve as reliable evaluators for text generation, with pairwise comparison being more reliable than direct scoring. Chen et al.~\citep{chen2024mllmjudge} extended this to the multimodal setting, showing that models like GPT-4V exhibit human-like discernment in pairwise visual comparisons. Our work applies this paradigm to a specialized scientific domain: rather than judging aesthetic quality or prompt adherence, we use multimodal LLMs to judge biological similarity between histopathology tiles, a task requiring domain-specific visual reasoning about cell morphology and tissue architecture.

\textbf{Generative Multimodal LLMs in Histopathology.}
Recent studies have begun to evaluate whether general-purpose multimodal LLMs can perform histopathology reasoning without task-specific training. Ferber et al.~\citep{ferber2024context} showed that GPT-4V with in-context examples can approach specialized models on selected histopathology classification tasks. However, large-scale pathology vision-language benchmarks report only moderate accuracy and suggest that models may exploit textual question or answer priors rather than consistently rely on visual evidence~\citep{gilal2025pathvlmeval}. These findings suggest that general-purpose MLLMs have visual reasoning capabilities but require stronger task structure for fine-grained histomorphologic interpretation. Our work shifts the focus from direct classification to probing whether MLLMs can make relational judgments of histological similarity, a task that provides stronger visual grounding by requiring explicit comparison between images rather than generating a class label from a single image.

\section{Method}
\label{sec:method}

\subsection{Relative Similarity Analysis}

We propose a relative similarity evaluation framework that tests whether vision models preserve correct tissue-type similarity structure in histopathology images (Figure~\ref{fig:schematic_pro}a). Given a set of tiles $\mathcal{T} = \{t_1, \ldots, t_N\}$ with associated class labels $y_i \in \mathcal{C}$ and a model-induced similarity function $s: \mathcal{T} \times \mathcal{T} \to \mathbb{R}$, we construct comparison tuples $(t_r, t_p, t_n)$ where the reference $t_r$ and positive $t_p$ share the same class ($y_r = y_p$) and the negative $t_n$ belongs to a different class ($y_n \neq y_r$). The model is evaluated on whether it correctly ranks the positive candidate as more similar to the reference than the negative:
\begin{equation}
\mathbbm{1}\bigl[s(t_r, t_p) > s(t_r, t_n)\bigr].
\end{equation}
Relative similarity accuracy, the proportion of correctly ordered comparisons, directly measures whether the model's similarity function respects the semantic class structure.

\subsubsection{Scope of Comparison}

\begin{figure}[!htb]
\centering
\includegraphics[width=\textwidth]{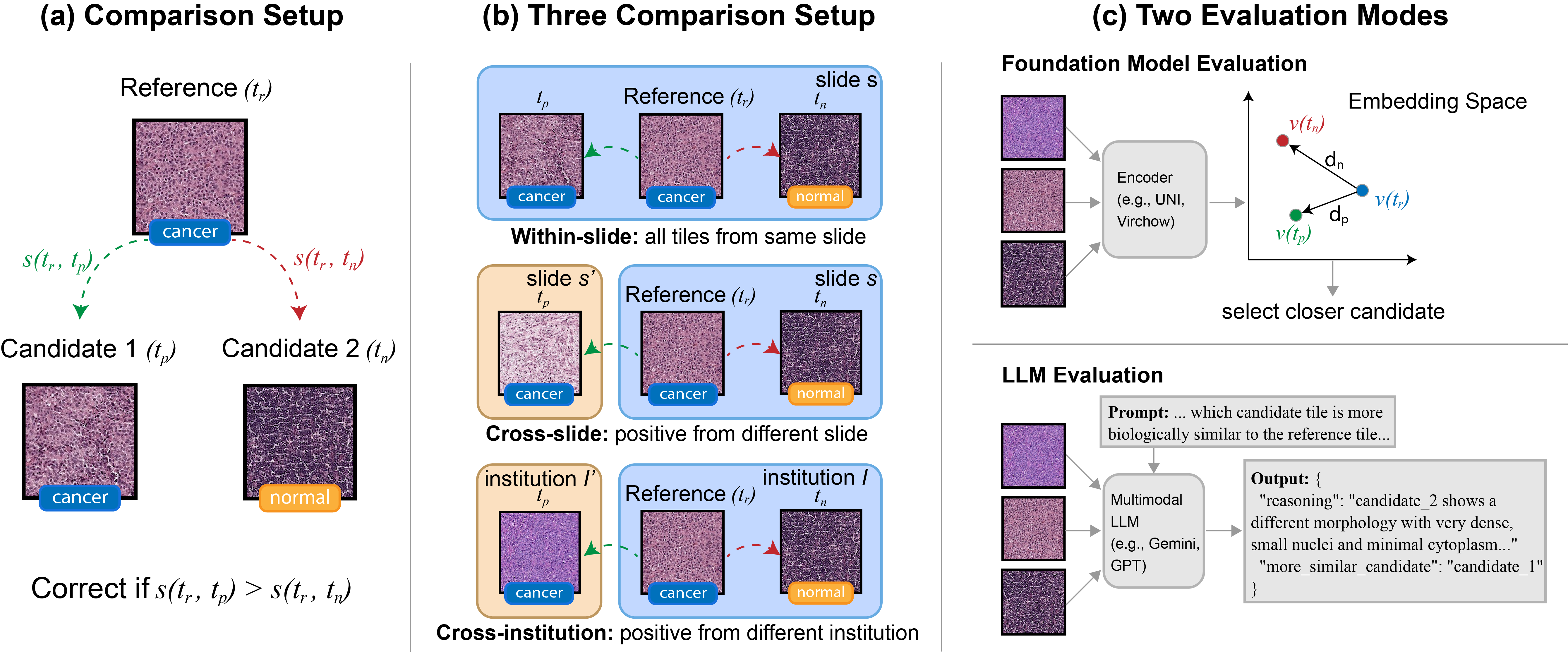}
\caption{Schematic of the relative similarity evaluation framework.}
\label{fig:schematic_pro}
\end{figure}

We define three comparison configurations (Figure~\ref{fig:schematic_pro}b) that introduce progressively stronger domain shifts, each testing a different aspect of the model's similarity structure.

\textbf{Within-slide.} All three tiles come from the same slide $s$: $t_r, t_p, t_n \in \mathcal{T}_s$. This tests whether the model can discriminate tissue classes in the absence of batch effects, providing an upper bound on class-discrimination ability.

\textbf{Cross-slide.} The reference $t_r$ and negative $t_n$ come from the same slide $s$, while the positive $t_p$ comes from a different slide $s'$ of the same class: $t_r, t_n \in \mathcal{T}_s$, $t_p \in \mathcal{T}_{s'}$, $s \neq s'$. The model must recognize that a tile from a different slide but the same tissue class is more similar to the reference than a tile from the same slide but a different class. This tests whether tissue-type similarity generalizes across slide-level variation.

\textbf{Cross-institution.} The reference $t_r$ and negative $t_n$ come from the same institution $I$, while the positive $t_p$ comes from a different institution $I'$: $t_r, t_n \in \mathcal{T}_I$, $t_p \in \mathcal{T}_{I'}$, $I \neq I'$. This is the most demanding configuration, testing whether the model can see past institution-level batch effects (scanner differences, staining protocols, tissue preparation) to identify tissue-type similarity.


\subsubsection{Model Evaluation}

\textbf{Foundation model evaluation.} For each comparison, we compute Euclidean distances between the reference embedding and each candidate embedding:
\begin{equation}
d_k = \| \mathbf{e}_\text{ref} - \mathbf{e}_{c_k} \|_2, \quad k \in \{1, 2\}.
\end{equation}
The model selects the candidate with smaller distance. This directly tests whether the embedding space assigns higher similarity to same-class tiles.

\textbf{LLM evaluation.} Each comparison is presented as three images (reference, candidate 1, candidate 2) along with a text prompt instructing the model to identify which candidate is more biologically similar to the reference. The prompt directs the model to focus on cell morphology (cell size, shape, nuclear features) and tissue architecture (glandular organization, stromal patterns, cell density), while ignoring non-biological differences such as staining intensity, imaging artifacts, or tissue preparation variations.

\textbf{Accuracy.} A comparison is scored as correct if the model selects the positive candidate. Comparisons where the model fails to return a valid response are excluded. Per-class accuracy aggregates over all comparisons where the reference shares a given class, and overall accuracy averages across all comparisons. We also compute per-class-pair accuracy matrices to identify which tissue type pairs are most easily confused.

\subsection{MOSAIC Benchmark}
\label{sec:benchmark}

We curate MOSAIC, a benchmark of six histopathology cohorts designed to evaluate cross-domain generalizability at two levels of domain shift. The datasets were selected to span (1) multiple organs (breast, colorectal, lung), (2) a range of tissue-class granularity (2--6 classes), (3) 9 distinct institutions across the cross-institution datasets, and (4) two distinct axes of domain shift: slide-level variation (different physical sections within a cohort) and institution-level variation (different scanners, staining protocols, and tissue preparation). HER2ST, CCTGS, and TIGER provide multi-slide cohorts for cross-slide evaluation, while CAMELYON16, TCGA, and BEETLE provide multi-institution cohorts for cross-institution evaluation. Together, these six datasets yield 13,800 comparison triplets, of which 8,400 are cross-domain (6,200 cross-slide and 2,200 cross-institution), that we release as a standardized evaluation resource alongside evaluation code and baseline model outputs.

The six component datasets are as follows. \underline{HER2ST}: breast cancer dataset with 8 slides (74,275 tiles), annotated with 5 tissue classes (adipose, breast glandular, cancer, connective, immune infiltrate)~\citep{andersson2021spatial}.
\underline{CCTGS}: 39 colorectal cancer slides (3,428,031 tiles) with 6 classes (adipose, lamina propria, lymphovascular invasion, muscularis propria, normal mucosa, tumor)~\citep{arslan2025colorectal}. \underline{TIGER}: 40 breast cancer slides (3,063,641 tiles) from the Tumor InfiltratinG lymphocytes in breast cancER grand-challenge with 4 classes (healthy glands, necrosis, stroma, tumor)~\citep{vanrijthoven2025tiger}. \underline{CAMELYON16}: 40 breast cancer slides (892,722 tiles) with 2 classes (tumor vs.\ normal) from 2 institutions (Radboud and Utrecht)~\citep{camelyon16}. \underline{TCGA}: 30 lung squamous cell carcinoma slides (601,155 tiles) from The Cancer Genome Atlas~\citep{tcga_lusc}, with annotations of invasive tumor~\citep{loeffler_2021_5320076}, yielding 2 classes (tumor vs.\ normal) across 3 tissue source sites (Mayo Clinic Rochester, the International Genomics Consortium, Indivumed). \underline{BEETLE}: 40 breast cancer slides (2,553,697 tiles) with 3 classes (invasive epithelium, necrosis, non-invasive epithelium) from 4 institutions (NKI, RUMC, SCH, TCGA)~\citep{lems2025multicentric}. For each dataset, WSIs are tiled using a sliding window (512$\times$512 pixels, stride 256) after Otsu-based tissue masking, and tiles are filtered by foreground proportion ($\geq$1\%).

For each cross-domain comparison triplet, a positive candidate shares the reference's tissue class but comes from a different slide or institution, while a negative candidate shares the reference's acquisition context but belongs to a different class. Candidates are randomly assigned to positions 1 and 2 with equal probability, so chance accuracy is 50\%. Triplet sampling details, including sampling strata and counts by dataset, are in Appendix~\ref{sec:experiment_setup}.

\subsection{Models}

We evaluate three categories of models. Pathology foundation models are domain-specific vision encoders pretrained on histopathology data: UNI2H~\citep{chen2024uni}, H-optimus-0~\citep{hoptimus0}, Midnight~\citep{KDK_Training_MICCAI2025}, CONCH v1.5~\citep{ding2024conchv15}, and Virchow2~\citep{zimmermann2024virchow2}. General foundation models include general-purpose vision encoders: DINOv2~\citep{oquab2023dinov2}, ResNet-50~\citep{he2016resnet}, Gemini Embedding 2~\citep{lee2025gemini}, and vision encoders extracted from multimodal LLM architectures (Qwen3.5-encoder~\citep{qwen3.5}, GLM-4.6V-encoder~\citep{vteam2025glm45}, Gemma 3-encoder~\citep{Kamath2025Gemma3T}). LLMs are multimodal LLMs used as zero-shot judges for relative similarity: Gemini 3 Flash~\citep{gemini3flash}, Gemini 3.1 Flash Lite~\citep{gemini31flashlite}, GPT-5 Nano~\citep{singh2025gpt5}, Gemma 3~\citep{Kamath2025Gemma3T}, GLM-4.6V~\citep{vteam2025glm45}, and Qwen3.5~\citep{qwen3.5}. All embedding extraction and local model inference were performed on an NVIDIA RTX 4090 GPU.

\section{Results}
\label{sec:results}

\begin{figure}[!htb]
\centering
\includegraphics[width=0.9\textwidth]{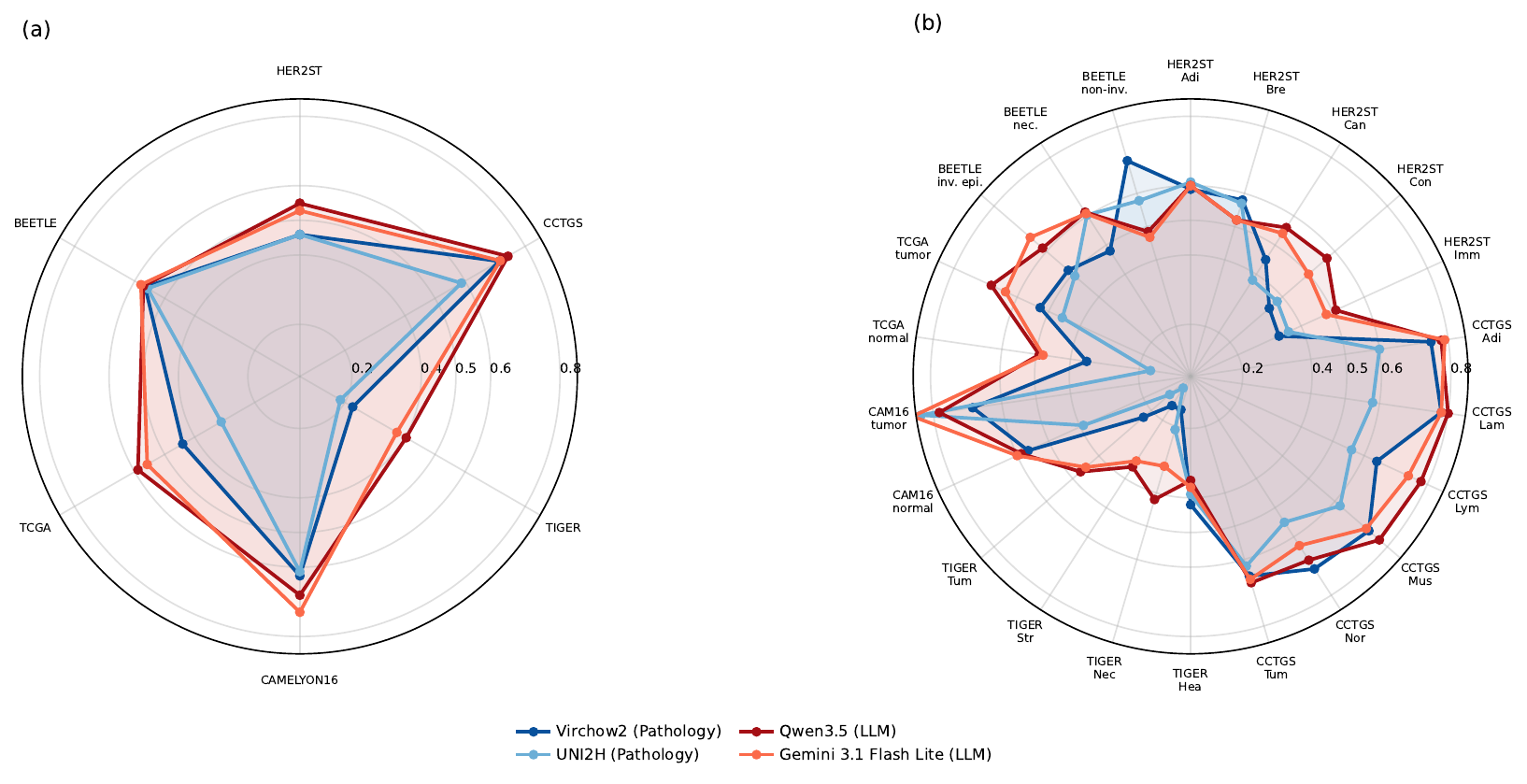}
\caption{Accuracy for relative similarity analysis across all the experiments.}
\label{fig:overall_radar}
\end{figure}

We test whether pathology foundation models preserve biological similarity when comparisons cross domain boundaries, using multimodal LLMs as a reference point for what is achievable without domain-specific training. \autoref{fig:overall_radar} summarizes results across all six datasets. The pattern is consistent: pathology foundation models perform well within a domain but degrade sharply across slides and institutions, while LLMs maintain robust performance. The advantage of LLMs is most pronounced in the hardest settings (cross-institution, morphologically similar classes) where foundation model shortcuts are most exposed.

\subsection{Cross-Slide Relative Similarity Analysis}

A robust computational pathology model should recognize the same tissue phenotype across different physical slides. However, slide-specific variation in tissue thickness, staining, and cutting artifacts can introduce technical signatures that models may exploit. We therefore evaluate cross-slide relative similarity on HER2ST, TIGER and CCTGS, testing whether models rank same-class tiles as more similar despite slide-level variation. Experiment setup details are in Appendix~\ref{sec:experiment_setup}.

As shown in Table~\ref{tab:cross_slide_merged}, multimodal LLMs outperform both pathology-specific and general-purpose foundation models in overall accuracy on all three datasets and achieve the best or second-best performance for most tissue classes. The LLM advantage over pathology foundation models is statistically significant on HER2ST (McNemar's $p < 0.0001$; Appendix~\ref{sec:statistical_significance}). These results demonstrate the potential ability of zero-shot multimodal LLMs to look past superficial slide-specific technical artifacts and match tissues based on genuine morphological features, such as cellular organization and structural architecture.


\begin{table}[!htb]
\centering
\caption{Cross-slide tile relative similarity accuracy on HER2ST, CCTGS, and TIGER, broken down by positive class $c_+$. For each column, the best method is in \textbf{bold} and the second best is \underline{underlined}. ``All'' refers to the overall accuracy across all classes. Class abbreviations are listed in Table~\ref{tab:class_abbreviations}.}
\label{tab:cross_slide_merged}
\resizebox{\textwidth}{!}{%
\begin{tabular}{l cccccc ccccccc ccccc}
\toprule
& \multicolumn{6}{c}{HER2ST} & \multicolumn{7}{c}{CCTGS} & \multicolumn{5}{c}{TIGER} \\
\cmidrule(lr){2-7} \cmidrule(lr){8-14} \cmidrule(lr){15-19}
Method & Adi & Bre & Can & Con & Imm & All & Adi & Lam & Lym & Mus & Nor & Tum & All & Hea & Nec & Str & Tum & All \\
\midrule
\multicolumn{19}{l}{\textit{Pathology foundation models}} \\
UNI2H & 0.61 & \underline{0.57} & 0.38 & 0.38 & 0.36 & 0.46 & 0.60 & 0.58 & 0.56 & 0.62 & 0.55 & 0.62 & 0.59 & 0.39 & 0.21 & 0.09 & 0.13 & 0.19 \\
H-optimus-0 & 0.59 & 0.53 & 0.36 & 0.40 & 0.32 & 0.44 & 0.59 & 0.56 & 0.57 & 0.59 & 0.51 & 0.51 & 0.55 & 0.40 & 0.20 & 0.17 & 0.17 & 0.22 \\
Midnight & 0.60 & 0.56 & 0.41 & 0.37 & 0.40 & 0.46 & 0.76 & 0.70 & 0.69 & 0.76 & \textbf{0.75} & \textbf{0.68} & 0.72 & \underline{0.50} & 0.12 & 0.11 & 0.16 & 0.20 \\
CONCH v1.5 & 0.43 & 0.52 & 0.51 & 0.41 & 0.38 & 0.45 & 0.47 & 0.59 & 0.58 & 0.66 & 0.53 & 0.55 & 0.56 & 0.32 & 0.19 & 0.18 & 0.20 & 0.21 \\
Virchow2 & 0.59 & \textbf{0.58} & 0.45 & 0.35 & 0.33 & 0.46 & 0.75 & 0.78 & 0.64 & 0.73 & \underline{0.71} & 0.65 & 0.71 & 0.42 & 0.15 & 0.15 & 0.23 & 0.23 \\
\midrule
\multicolumn{19}{l}{\textit{General foundation models}} \\
Gemini Emb.\ 2 & 0.54 & 0.50 & 0.51 & 0.45 & 0.43 & 0.48 & 0.67 & 0.77 & 0.69 & 0.69 & 0.63 & 0.61 & 0.67 & 0.43 & 0.23 & 0.35 & 0.40 & 0.36 \\
DINOv2 & 0.61 & 0.51 & \underline{0.57} & 0.49 & \underline{0.56} & \underline{0.55} & 0.50 & 0.63 & 0.66 & 0.65 & 0.60 & 0.61 & 0.60 & 0.36 & 0.17 & \textbf{0.45} & 0.39 & 0.36 \\
ResNet-50 & 0.55 & 0.54 & 0.45 & 0.33 & 0.41 & 0.45 & 0.49 & 0.70 & 0.66 & 0.65 & 0.62 & 0.60 & 0.61 & 0.28 & 0.16 & 0.36 & 0.34 & 0.30 \\
Qwen3.5-enc. & \underline{0.63} & 0.50 & 0.38 & 0.32 & 0.36 & 0.44 & 0.59 & 0.61 & 0.56 & 0.56 & 0.55 & 0.54 & 0.57 & 0.41 & \underline{0.52} & 0.29 & 0.30 & 0.36 \\
GLM-4.6V-enc. & 0.57 & 0.49 & 0.44 & 0.41 & 0.46 & 0.48 & 0.53 & 0.61 & 0.60 & 0.63 & 0.62 & 0.57 & 0.59 & 0.31 & 0.33 & 0.36 & 0.34 & 0.34 \\
Gemma 3-enc. & 0.54 & 0.49 & 0.40 & 0.31 & 0.34 & 0.41 & 0.63 & 0.75 & 0.72 & 0.66 & 0.60 & 0.60 & 0.65 & 0.40 & 0.19 & 0.29 & 0.45 & 0.34 \\
\midrule
\multicolumn{19}{l}{\textit{LLMs}} \\
Gemini 3 Flash & 0.61 & 0.53 & 0.55 & \underline{0.54} & 0.53 & \underline{0.55} & \textbf{0.79} & \underline{0.79} & \textbf{0.80} & \textbf{0.78} & 0.68 & \underline{0.67} & \textbf{0.75} & 0.35 & 0.28 & 0.31 & 0.44 & 0.35 \\
Gemini 3.1 Flash Lite & 0.60 & 0.52 & 0.54 & 0.50 & 0.48 & 0.53 & \textbf{0.79} & 0.78 & 0.74 & 0.72 & 0.63 & 0.66 & 0.72 & 0.37 & 0.32 & 0.34 & 0.45 & 0.37 \\
Qwen3.5 & 0.60 & 0.52 & 0.56 & \textbf{0.57} & 0.51 & \underline{0.55} & 0.78 & \textbf{0.80} & \underline{0.78} & \underline{0.77} & 0.68 & \underline{0.67} & \underline{0.74} & 0.35 & 0.42 & 0.36 & \underline{0.47} & 0.40 \\
GLM-4.6V & 0.59 & 0.50 & \underline{0.57} & 0.51 & \textbf{0.58} & \underline{0.55} & 0.64 & 0.73 & 0.73 & 0.76 & 0.64 & 0.64 & 0.68 & 0.48 & 0.31 & 0.38 & 0.40 & 0.39 \\
Gemma 3 & 0.55 & 0.54 & \textbf{0.58} & 0.50 & 0.52 & 0.54 & 0.65 & 0.67 & 0.71 & 0.71 & 0.62 & 0.56 & 0.65 & 0.41 & 0.36 & \underline{0.41} & \textbf{0.50} & \underline{0.43} \\
GPT-5 Nano & \textbf{0.64} & \textbf{0.58} & 0.56 & 0.50 & \underline{0.56} & \textbf{0.57} & 0.60 & 0.68 & 0.66 & 0.73 & 0.62 & 0.60 & 0.64 & \textbf{0.56} & \textbf{0.56} & 0.39 & 0.41 & \textbf{0.46} \\
\bottomrule
\end{tabular}%
}
\end{table}

\subsubsection{Class-Level Confusion}

To better characterize the failure modes behind this performance gap, we analyze which tissue classes are most frequently conflated in cross-slide comparisons. Per-class-pair confusion heatmaps (Appendix Figures~\ref{fig:cross_slide_perclass},~\ref{fig:cross_slide_perclass_cctgs}, and~\ref{fig:cross_slide_perclass_tiger}) show that foundation models most easily confuse morphologically similar classes. On HER2ST, for example, UNI2H achieves only 25\% accuracy on the connective-vs-immune-infiltrate pair and 32\% on the cancer-vs-immune-infiltrate pair, both well below chance. MLLMs improve most on these difficult pairs, where foundation models fall below chance, rather than uniformly improving all class pairs. Across all datasets, LLMs also show more uniform per-class accuracy: UNI2H accuracy on TIGER ranges from 0.09 to 0.39 across classes (spread of 0.30), while GPT-5 Nano ranges from 0.39 to 0.56 (spread of 0.17).

\subsubsection{Comparison with Within-Slide Performance}

\begin{figure}[!htb]
\centering
\includegraphics[width=\textwidth]{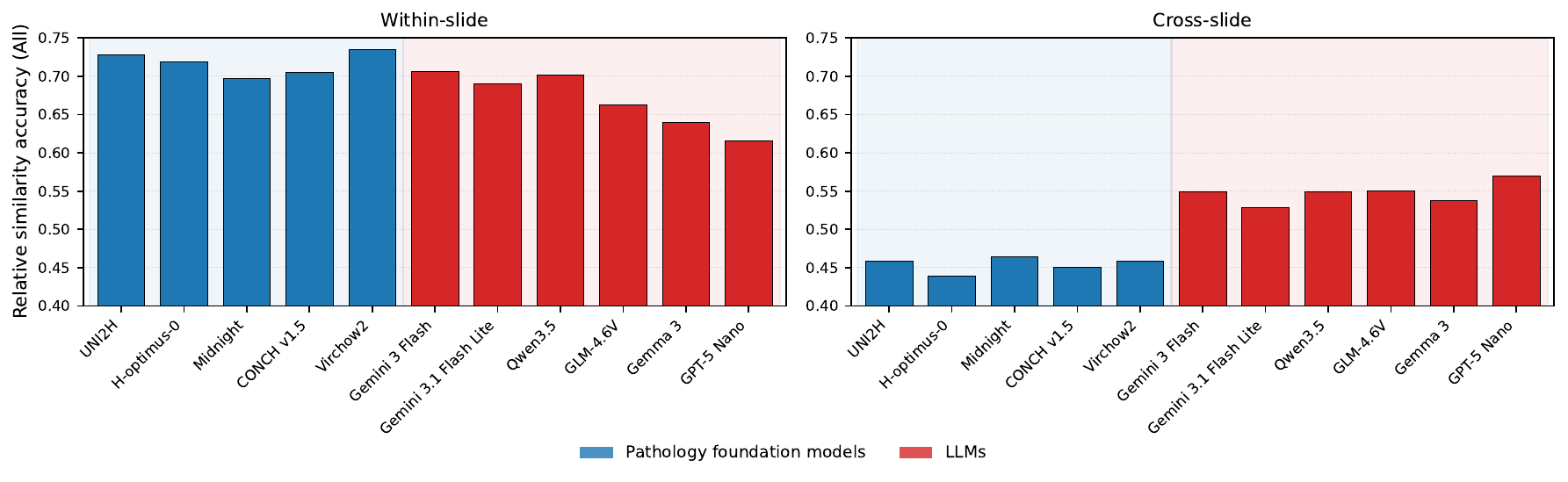}
\caption{Within-slide (left) and cross-slide (right) relative similarity accuracy on HER2ST. Pathology foundation models lead within-slide accuracy, while LLMs achieve competitive or higher cross-slide accuracy.}
\label{fig:within_vs_cross_bar}
\end{figure}

The cross-slide results above show that foundation models struggle when comparisons cross slide boundaries. To disentangle the effect of domain shift from inherent class-discrimination limitations, we introduce a within-slide baseline on HER2ST, where the reference, positive, and negative tiles all come from the same slide ($t_r, t_p, t_n \in \mathcal{T}_s$). We then compare this baseline with cross-slide accuracy, where the positive tile comes from a different slide ($t_p \in \mathcal{T}_{s'}$, $s' \neq s$; \autoref{fig:within_vs_cross_bar}). In the within-slide setting, pathology foundation models dominate: Virchow2 leads at 73.5\%, followed by UNI2H (72.8\%). When comparisons cross slide boundaries, all models drop in accuracy, but pathology foundation models suffer the largest decline while LLMs maintain relatively higher cross-slide accuracy. This suggests that foundation models capture slide-specific features well but struggle to generalize across slides, whereas LLMs trade some within-slide precision for better cross-slide robustness. Within-slide accuracy breakdown is in Appendix~\ref{sec:within_slide_appendix} (Table~\ref{tab:within_slide_acc}); scatter plot comparison is in Figure~\ref{fig:within_vs_cross_merged_scatter}.

\subsection{Cross-Institution Relative Similarity Analysis}

The cross-slide analysis shows that slide-level variation already degrades foundation model performance. Cross-institution comparisons introduce a stronger domain shift, where the positive $t_p$ comes from a different institution $I'$ with different scanners, staining protocols, and tissue preparation, while the reference $t_r$ and negative $t_n$ come from the same institution $I$. We evaluate cross-institutional relative similarity on CAMELYON16, TCGA, and BEETLE. Experiment setup details are in Appendix~\ref{sec:experiment_setup}.

\begin{figure}[!htb]
\centering
\includegraphics[width=0.8\textwidth]{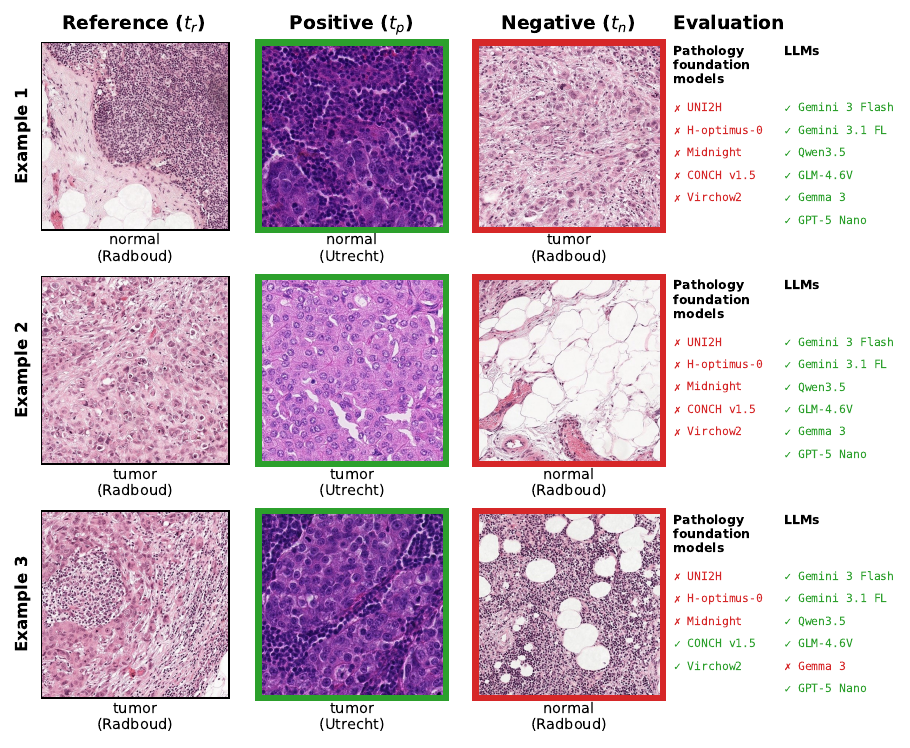}
\caption{Example cross-institution comparisons on CAMELYON16. Each row shows one comparison: reference $t_r$ (left), positive $t_p$ (center, green border, same class from a different institution), negative $t_n$ (right, red border, different class from the same institution). Correctness of each model's choice is shown on the right (\checkmark\ correct, $\times$ incorrect).}
\label{fig:examples}
\end{figure}

Table~\ref{tab:cross_inst_all} reports overall accuracy across all three datasets. LLMs generally outperform foundation models, with the advantage most pronounced on CAMELYON16 (Gemini 3.1 Flash Lite: 73.0\% vs best foundation model Midnight: 64.0\%, McNemar's $p < 0.001$; Appendix~\ref{sec:statistical_significance}) and TCGA (Qwen3.5: 58.9\% vs best foundation model Gemini Emb.\ 2: 50.3\%). Pathology foundation models show high variance across institutions and perform worst on TCGA (H-optimus-0: 17.7\%), indicating that institution-level batch effects dominate their similarity structure. When the reference and positive candidate are normal tiles, most foundation models fall below chance across datasets, while LLMs more frequently maintain above-chance accuracy. A qualitative examination of these failures makes the mechanics of this shortcut learning visually apparent (\autoref{fig:examples}). Foundation model embeddings often favor color-matched negative tiles from the same institution, whereas MLLMs more often recover the biological match despite staining differences. Per-institution breakdowns are in Appendix~\ref{sec:cross_inst_appendix} (Tables~\ref{tab:cross_inst_camelyon16}, \ref{tab:cross_inst_tcga}, and~\ref{tab:cross_inst_beetle}).

\begin{table}[!htb]
\centering
\caption{Cross-institution relative similarity accuracy on CAMELYON16 (2 institutions, 2 classes), TCGA (3 institutions, 2 classes), and BEETLE (4 institutions, 3 classes). For each column, the best method is in \textbf{bold} and the second best is \underline{underlined}. Per-institution breakdowns are in Appendix~\ref{sec:cross_inst_appendix}.}
\label{tab:cross_inst_all}
\footnotesize
\begin{tabular}{l ccc}
\toprule
Method & CAMELYON16 & TCGA & BEETLE \\
\midrule
\multicolumn{4}{l}{\textit{Pathology foundation models}} \\
UNI2H & 0.61 & 0.31 & 0.56 \\
H-optimus-0 & 0.40 & 0.18 & 0.45 \\
Midnight & 0.64 & 0.36 & 0.48 \\
CONCH v1.5 & 0.57 & 0.37 & 0.47 \\
Virchow2 & 0.63 & 0.44 & \underline{0.57} \\
\midrule
\multicolumn{4}{l}{\textit{General foundation models}} \\
Gemini Emb.\ 2 & 0.61 & 0.50 & 0.56 \\
DINOv2 & 0.52 & 0.44 & 0.51 \\
ResNet-50 & 0.49 & 0.44 & 0.46 \\
Qwen3.5-enc. & 0.56 & 0.41 & 0.33 \\
GLM-4.6V-enc. & 0.58 & 0.40 & 0.48 \\
Gemma 3-enc. & 0.60 & 0.39 & 0.55 \\
\midrule
\multicolumn{4}{l}{\textit{LLMs}} \\
Gemini 3 Flash & \underline{0.71} & 0.55 & 0.54 \\
Gemini 3.1 Flash Lite & \textbf{0.73} & \underline{0.56} & \textbf{0.58} \\
Qwen3.5 & 0.68 & \textbf{0.59} & \underline{0.57} \\
GLM-4.6V & 0.61 & 0.47 & 0.50 \\
Gemma 3 & 0.56 & 0.53 & 0.56 \\
GPT-5 Nano & 0.59 & 0.52 & \underline{0.57} \\
\bottomrule
\end{tabular}
\end{table}


\subsubsection{Comparison with Within-Institution Performance}

\begin{figure}[!htb]
\centering
\includegraphics[width=0.9\textwidth]{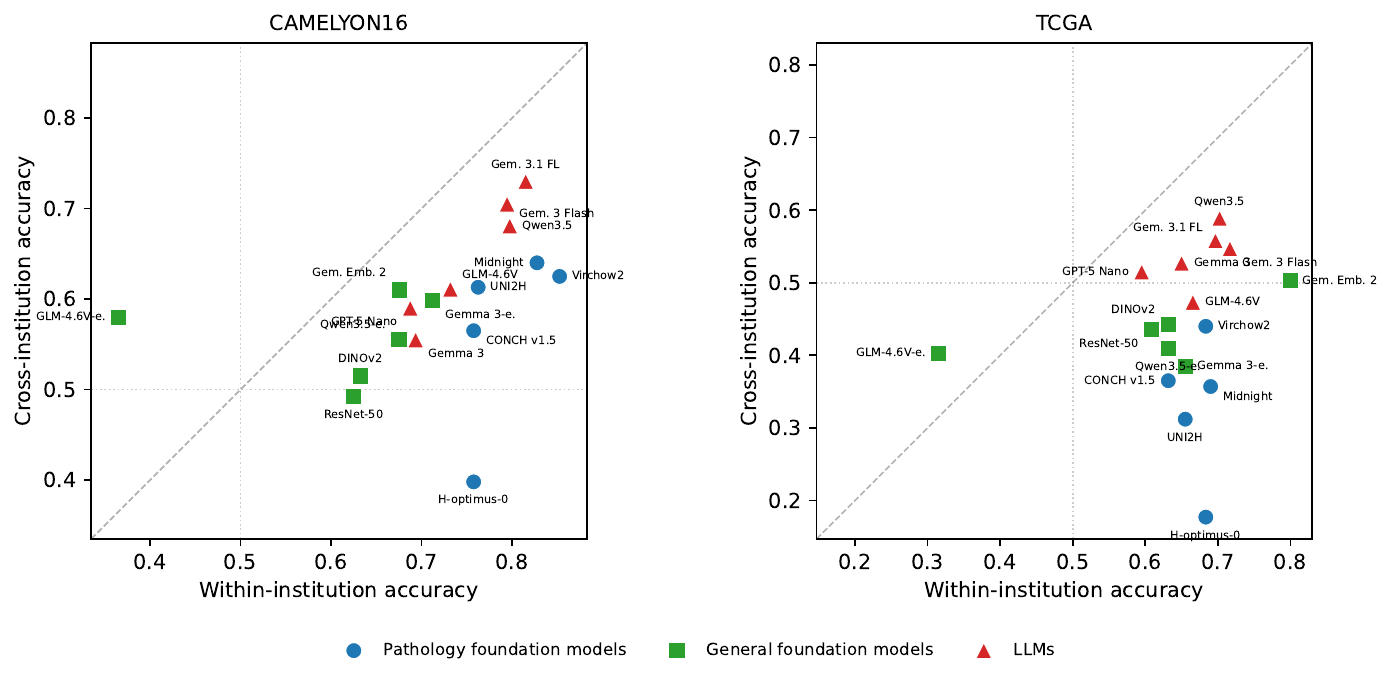}
\caption{Within-institution vs cross-institution relative similarity accuracy on CAMELYON16 (left) and TCGA (right). Each point is a model.}
\label{fig:within_vs_cross_institution_scatter}
\end{figure}

As with the cross-slide analysis, we compare within-domain and cross-domain accuracy to quantify the effect of institution-level domain shift. Figure~\ref{fig:within_vs_cross_institution_scatter} compares within-institution accuracy (where $t_r, t_p, t_n \in \mathcal{T}_I$) with cross-institution accuracy (where $t_p \in \mathcal{T}_{I'}$, $I' \neq I$). On both CAMELYON16 and TCGA, all models fall below the diagonal, confirming that cross-institution discrimination is harder. The gap is larger on TCGA, where pathology foundation models cluster in the bottom-right with high within-institution accuracy (63--69\%) but low cross-institution accuracy (18--44\%). LLMs maintain higher cross-institution accuracy relative to their within-institution performance, mirroring the within-slide vs cross-slide pattern observed in Figure~\ref{fig:within_vs_cross_bar}. This consistent pattern across both slide-level and institution-level domain shifts suggests that foundation models encode domain-specific features well but struggle to generalize, whereas LLMs are more robust to batch effects. Per-institution within-institution breakdowns are in Appendix~\ref{sec:within_inst_supplementary}.

\subsection{Scaling Training Data Does Not Resolve the Robustness Gap}

\begin{figure}[!htb]
\centering
\includegraphics[width=\textwidth]{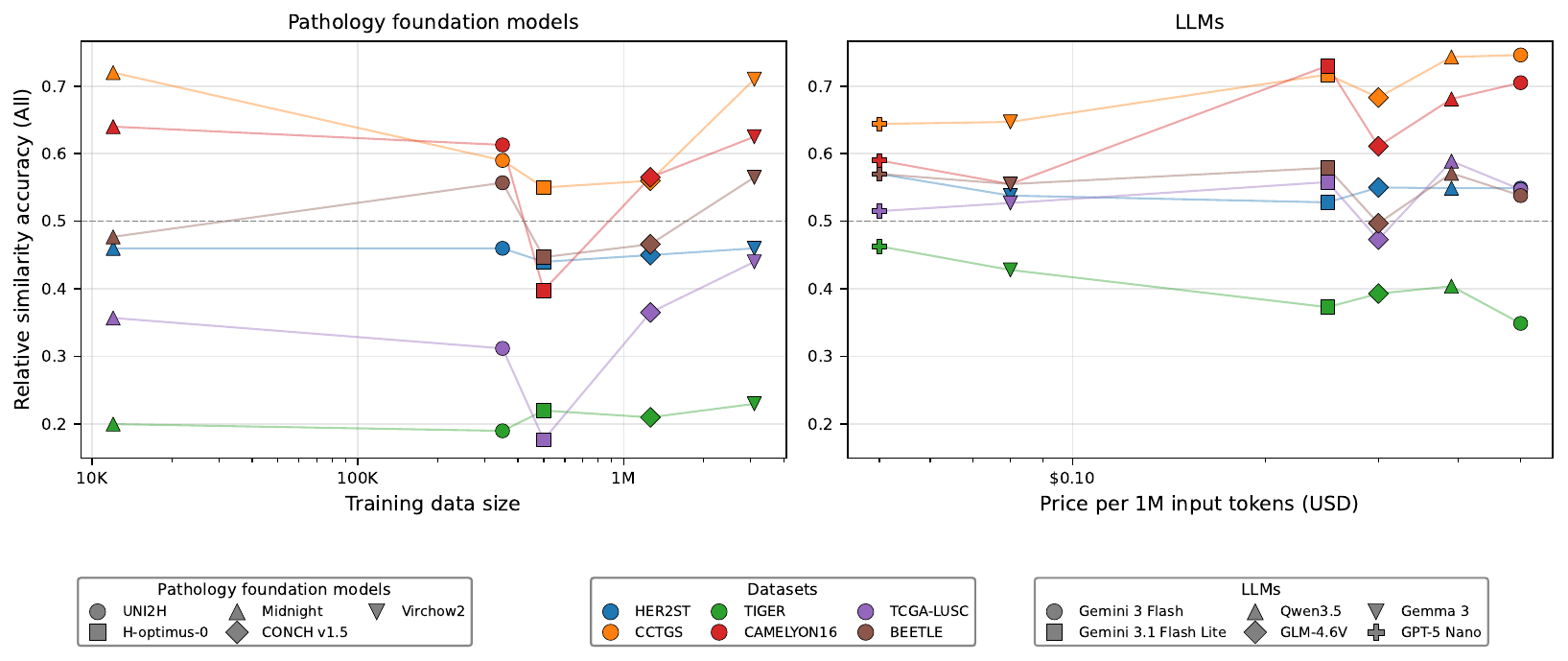}
\caption{Relative similarity accuracy vs model cost/size across all six datasets.}
\label{fig:cost_accuracy_llm_single}
\end{figure}

A natural hypothesis is that the cross-domain failures of pathology foundation models reflect insufficient training data, and that models trained on more tiles would learn to separate biology from batch effects. We test this directly. Figure~\ref{fig:cost_accuracy_llm_single} shows relative similarity accuracy across all six datasets as a function of training data size (for pathology models) and input-token price as a proxy for deployment cost (for LLMs). Within the evaluated pathology encoders, larger reported training sets do not translate into better cross-domain similarity accuracy: models trained on over 100 million tiles perform comparably to those trained on far fewer. This suggests that training-set scale alone is insufficient and points to the learning objective as a likely driver. For LLMs, even the cheapest models consistently outperform pathology foundation models, indicating that the advantage is not driven by model scale or cost. General foundation models and per-dataset LLM cost-accuracy panels are in Appendix~\ref{sec:cost_accuracy_appendix}. Training data sizes are listed in Appendix~\ref{sec:training_data_sources} (Table~\ref{tab:training_data_sources}).

\subsection{Ablation Studies}

To assess the contribution of the biology-focused prompt, we compare it against a minimal prompt that simply asks which candidate is more similar to the reference without any domain-specific guidance (full prompt text in Appendix~\ref{sec:prompt_ablation_appendix}). The biology-focused prompt instructs the LLM to focus on cell morphology and tissue architecture while ignoring non-biological differences such as staining intensity and imaging artifacts. Table~\ref{tab:ablation} reports accuracy for Gemini 3 Flash and GPT-5 Nano on HER2ST (cross-slide) and CAMELYON16 (cross-institution). For both models and both datasets, the biology-focused prompt yields higher accuracy than the minimal prompt. Even the minimal prompt outperforms pathology foundation models on CAMELYON16 (0.668 vs best FM 0.640). This highlights a practical advantage of LLMs: similarity criteria can be steered at inference time, whereas foundation model embeddings are fixed after pretraining.

\begin{table}[htbp]
\centering
\caption{Effect of LLM prompt on relative similarity accuracy.}
\label{tab:ablation}
\footnotesize
\begin{tabular}{l cc cc}
\toprule
& \multicolumn{2}{c}{HER2ST (cross-slide)} & \multicolumn{2}{c}{CAMELYON16 (cross-inst.)} \\
\cmidrule(lr){2-3} \cmidrule(lr){4-5}
Prompt & Gemini 3 Flash & GPT-5 Nano & Gemini 3 Flash & GPT-5 Nano \\
\midrule
Biology-focused (default) & 0.549 & 0.570 & 0.705 & 0.590 \\
Minimal & 0.510 & 0.560 & 0.668 & 0.555 \\
\bottomrule
\end{tabular}
\end{table}

\section{Conclusion}
\label{sec:conclusion}

Pathology foundation models can appear robust under standard within-domain evaluations while failing to preserve biological similarity across institutions, a clinically consequential gap that scaling training data alone does not resolve. Multimodal LLMs, without domain-specific training, consistently outperform these specialized models in cross-domain settings, demonstrating that robust histological similarity is achievable today and highlighting the potential of LLMs as a tool for mitigating batch effects in computational pathology. We release MOSAIC, a cross-domain benchmark spanning 6 histopathology cohorts, to support future development of models that prioritize biological signal over acquisition-context shortcuts. Our evaluation is limited to H\&E-stained images. Future work could extend this framework to other pathology imaging modalities such as immunohistochemistry and multiplexed imaging, as well as spatial biology data where cross-platform batch effects pose similar challenges.

\bibliographystyle{plainnat}
\bibliography{references}

\appendix
\newpage

\section{Experiment Setup}
\label{sec:experiment_setup}

\subsection{Within-Slide Comparisons}

All three tiles come from the same slide $s$: $t_r, t_p, t_n \in \mathcal{T}_s$, with $t_r$ and $t_p$ sharing class $c_+$ and $t_n$ from class $c_-$. For HER2ST, 20 comparisons are sampled per (slide, $c_+$, $c_-$) combination. The positive and negative are randomly assigned to positions 1 and 2 with equal probability, so chance accuracy is 50\%.

\subsection{Cross-Slide Comparisons}

For each ordered class pair $(c_+, c_-)$ with $c_+ \neq c_-$, we sample 100 comparisons where the reference $t_r$ has class $c_+$, the positive $t_p \in \mathcal{T}_{s'}$ has class $c_+$ from a different slide ($s' \neq s$), and the negative $t_n \in \mathcal{T}_s$ has class $c_-$ from the same slide as the reference. HER2ST (5 classes) yields 20 ordered pairs for 2000 total comparisons. CCTGS (6 classes, 39 slides) yields 30 pairs for 3000 total, and TIGER (4 classes, 40 slides) yields 12 pairs for 1200 total. The positive and negative are randomly assigned to positions 1 and 2 with equal probability, so chance accuracy is 50\%.

\subsection{Within-Institution Comparisons}

All three tiles come from the same institution $I$: $t_r, t_p, t_n \in \mathcal{T}_I$, with $t_r$ and $t_p$ sharing class $c_+$ and $t_n$ from class $c_-$. Tiles may come from different slides within the institution. For each (institution, $c_+$, $c_-$) combination, 100 comparisons are sampled.

\subsection{Cross-Institution Comparisons}

For each (institution $I$, $c_+$) combination, 100 comparisons are sampled where the reference $t_r \in \mathcal{T}_I$ has class $c_+$, the positive $t_p \in \mathcal{T}_{I'}$ has class $c_+$ from a different institution ($I' \neq I$), and the negative $t_n \in \mathcal{T}_I$ has class $c_- \neq c_+$ from the same institution (with $c_-$ drawn uniformly from available classes). CAMELYON16 has 2 institutions (Radboud with 21 slides, Utrecht with 19 slides) and 2 classes (tumor vs normal, assigned by computing tumor fraction from pixel-level masks with threshold $>$50\%), yielding 4 combinations and 400 total comparisons. TCGA has 3 tissue source sites (TSS-22, TSS-56, TSS-66, each with 10 slides) and 2 classes, yielding 6 combinations and 600 total. BEETLE has 4 institutions (NKI, RUMC, SCH, TCGA, each with 10 slides) and 3 classes (invasive epithelium, necrosis, non-invasive epithelium derived from pixel-level segmentation masks), yielding 12 combinations and 1200 total.

\clearpage
\section{Cross-Slide Supplementary Results}
\label{sec:cross_slide_supplementary}

\subsection{Within-Slide Accuracy}
\label{sec:within_slide_appendix}

\begin{table}[!htb]
\centering
\caption{Within-slide tile relative similarity accuracy on HER2ST, broken down by tissue class.}
\label{tab:within_slide_acc}
\resizebox{\textwidth}{!}{%
\begin{tabular}{lcccccc}
\toprule
Method & Adipose & Breast gl. & Cancer & Connective & Immune inf. & All \\
\midrule
\multicolumn{7}{l}{\textit{Pathology foundation models}} \\
UNI2H & \textbf{0.775} & \textbf{0.759} & 0.664 & \textbf{0.757} & 0.712 & 0.728 \\
H-optimus-0 & 0.709 & 0.691 & 0.682 & 0.748 & \textbf{0.756} & 0.719 \\
Midnight & 0.706 & 0.732 & 0.670 & 0.695 & 0.700 & 0.697 \\
CONCH v1.5 & 0.706 & 0.736 & 0.691 & 0.689 & 0.724 & 0.705 \\
Virchow2 & 0.741 & 0.718 & \textbf{0.755} & 0.739 & 0.709 & \textbf{0.735} \\
\midrule
\multicolumn{7}{l}{\textit{General foundation models}} \\
Gemini Emb.\ 2 & 0.616 & 0.591 & 0.720 & 0.691 & 0.641 & 0.662 \\
DINOv2 & 0.681 & 0.523 & 0.689 & 0.566 & 0.674 & 0.633 \\
ResNet-50 & 0.675 & 0.691 & 0.689 & 0.700 & 0.676 & 0.687 \\
Qwen3.5-enc. & 0.734 & 0.605 & 0.650 & 0.591 & 0.647 & 0.644 \\
GLM-4.6V-enc. & 0.662 & 0.582 & 0.670 & 0.623 & 0.671 & 0.646 \\
Gemma 3-enc. & 0.641 & 0.614 & 0.720 & 0.632 & 0.688 & 0.664 \\
\midrule
\multicolumn{7}{l}{\textit{LLMs}} \\
Gemini 3 Flash & 0.728 & 0.659 & 0.716 & 0.718 & 0.685 & 0.706 \\
Gemini 3.1 Flash Lite & 0.744 & 0.609 & 0.707 & 0.727 & 0.621 & 0.690 \\
Qwen3.5 & 0.713 & 0.623 & 0.705 & 0.745 & 0.682 & 0.702 \\
GLM-4.6V & 0.684 & 0.568 & 0.675 & 0.671 & 0.676 & 0.663 \\
Gemma 3 & 0.697 & 0.600 & 0.691 & 0.609 & 0.588 & 0.640 \\
GPT-5 Nano & 0.669 & 0.573 & 0.634 & 0.577 & 0.624 & 0.616 \\
\bottomrule
\end{tabular}%
}
\end{table}

\begin{figure}[!htb]
\centering
\includegraphics[width=0.45\textwidth]{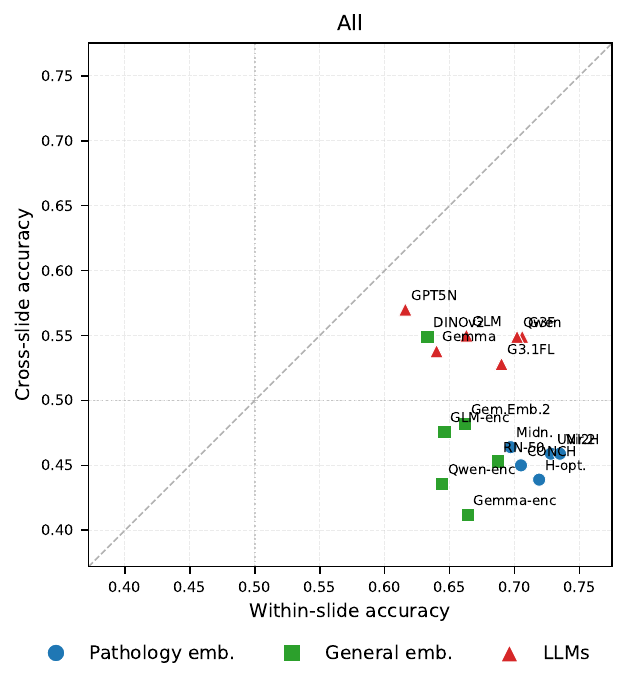}
\caption{Within-slide vs cross-slide relative similarity accuracy on HER2ST (overall accuracy). Each point is a method, colored/marked by category (pathology foundation models = blue circles, general foundation models = green squares, LLMs = red triangles) and labeled with an abbreviated name. Dashed diagonal: $y=x$; dotted gray lines: chance (0.5). All methods fall below the diagonal, confirming that cross-slide discrimination is harder. Pathology foundation models cluster in the lower-right (high within-slide, low cross-slide), while LLMs maintain higher cross-slide accuracy relative to their within-slide performance.}
\label{fig:within_vs_cross_merged_scatter}
\end{figure}

\subsection{Per-Class Pair Confusion Heatmaps}
\label{sec:perclass_heatmaps_appendix}

\begin{figure}[!htb]
\centering
\includegraphics[width=\textwidth]{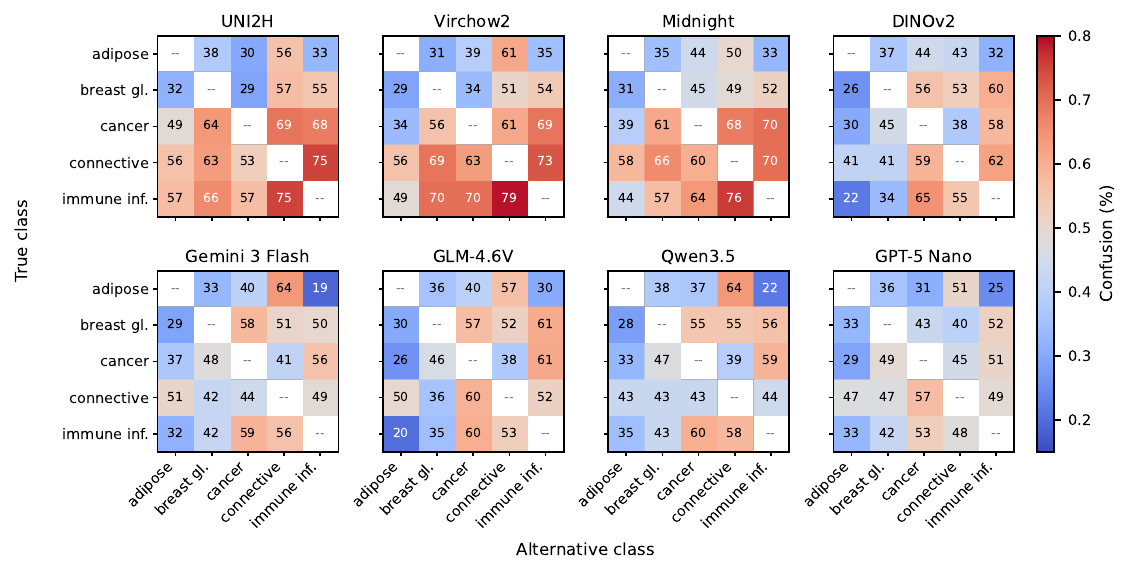}
\caption{Per-class pair confusion for cross-slide comparisons on HER2ST. Rows indicate the positive class $c_+$ (shared by reference $t_r$ and positive $t_p$) and columns the negative class $c_-$. Higher values (warmer colors) indicate greater confusion.}
\label{fig:cross_slide_perclass}
\end{figure}

\begin{figure}[!htb]
\centering
\includegraphics[width=\textwidth]{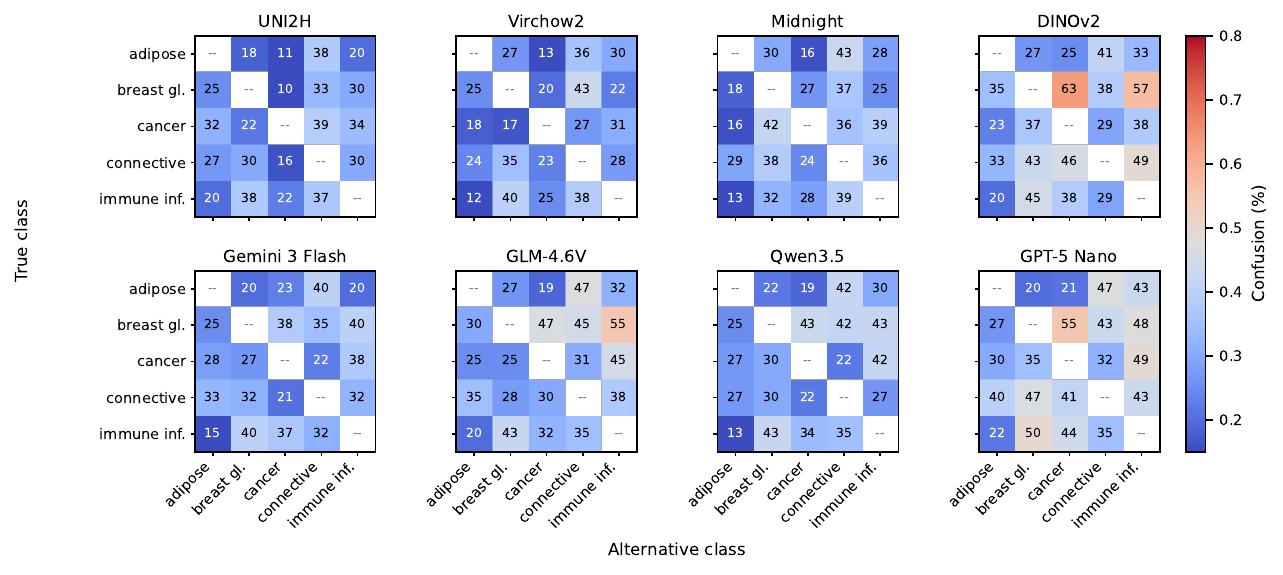}
\caption{Per-class pair confusion for within-slide comparisons on HER2ST. Rows indicate the positive class $c_+$ and columns the negative class $c_-$. Higher values (warmer colors) indicate greater confusion.}
\label{fig:within_slide_perclass}
\end{figure}

\begin{figure}[!htb]
\centering
\includegraphics[width=\textwidth]{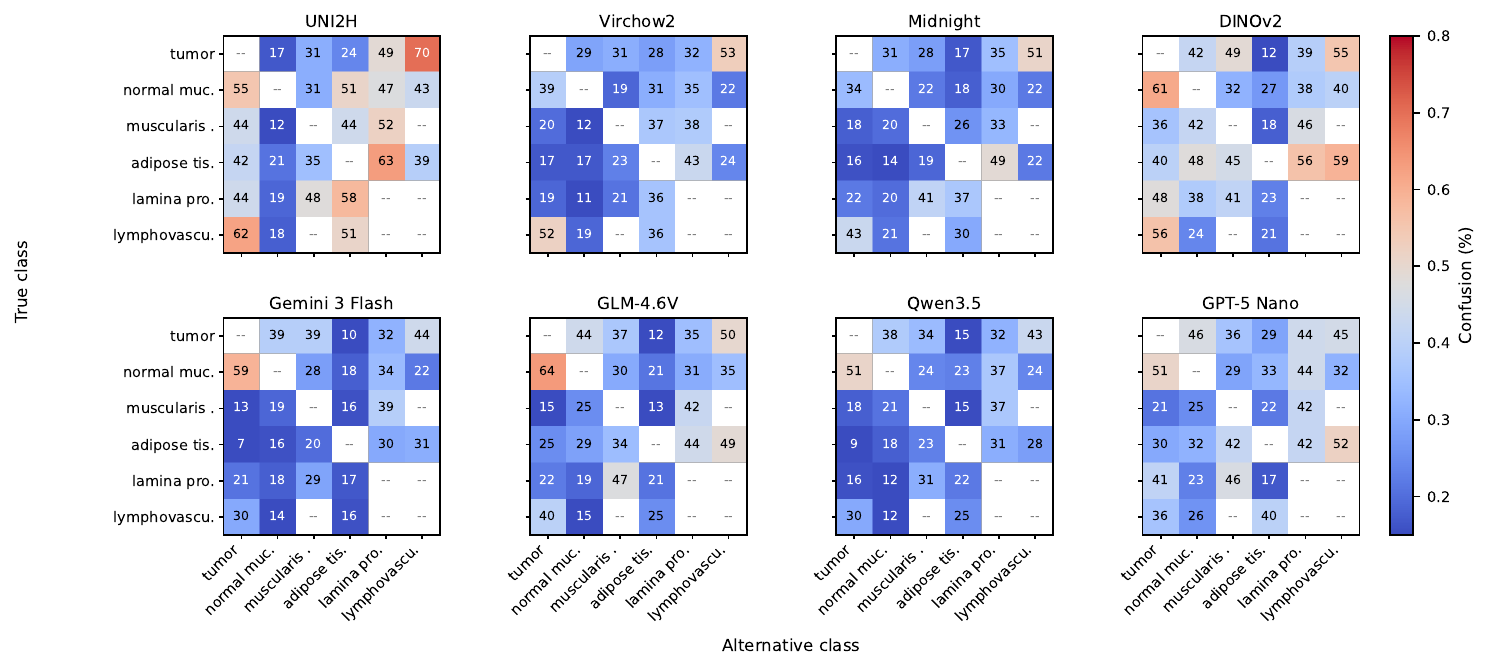}
\caption{Per-class pair confusion for cross-slide comparisons on CCTGS. Rows indicate the positive class $c_+$ and columns the negative class $c_-$. Higher values (warmer colors) indicate greater confusion.}
\label{fig:cross_slide_perclass_cctgs}
\end{figure}

\begin{figure}[!htb]
\centering
\includegraphics[width=\textwidth]{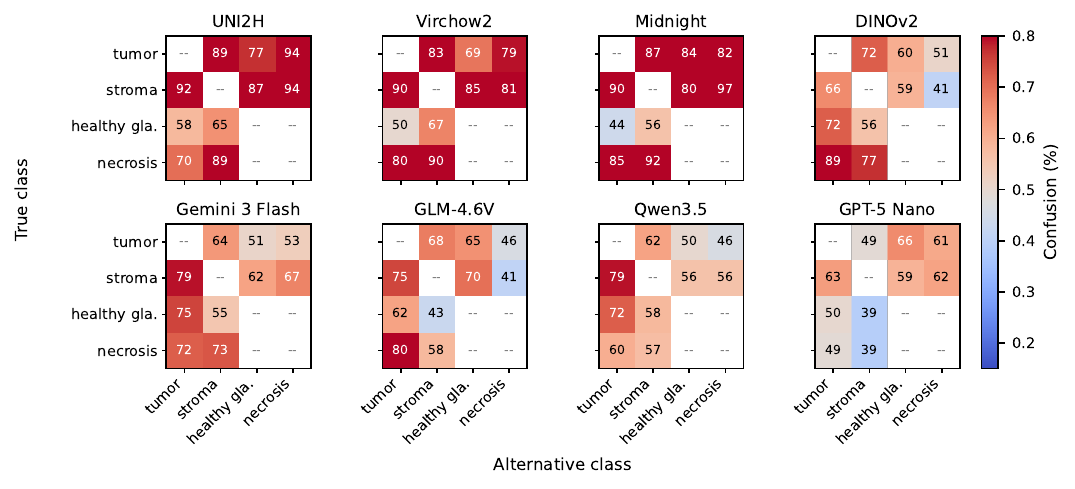}
\caption{Per-class pair confusion for cross-slide comparisons on TIGER. Rows indicate the positive class $c_+$ and columns the negative class $c_-$. Higher values (warmer colors) indicate greater confusion.}
\label{fig:cross_slide_perclass_tiger}
\end{figure}

\clearpage
\section{Cross-Institution Supplementary Results}
\label{sec:cross_inst_supplementary}

\subsection{Per-Institution Breakdown}
\label{sec:cross_inst_appendix}

\begin{table}[!htb]
\centering
\caption{Cross-institution tile relative similarity accuracy on CAMELYON16.}
\label{tab:cross_inst_camelyon16}
\resizebox{\textwidth}{!}{%
\begin{tabular}{lccccc}
\toprule
Method & Radboud normal & Radboud tumor & Utrecht normal & Utrecht tumor & All \\
\midrule
\multicolumn{6}{l}{\textit{Pathology foundation models}} \\
UNI2H & 0.340 & 0.770 & 0.440 & 0.900 & 0.613 \\
H-optimus-0 & 0.390 & 0.310 & 0.490 & 0.400 & 0.398 \\
Midnight & \textbf{0.610} & 0.670 & 0.550 & 0.730 & 0.640 \\
CONCH v1.5 & 0.460 & 0.630 & 0.410 & 0.760 & 0.565 \\
Virchow2 & 0.500 & 0.660 & \textbf{0.630} & 0.710 & 0.625 \\
\midrule
\multicolumn{6}{l}{\textit{General foundation models}} \\
Gemini Emb.\ 2 & 0.510 & 0.800 & 0.400 & 0.730 & 0.610 \\
DINOv2 & 0.370 & 0.750 & 0.250 & 0.690 & 0.515 \\
ResNet-50 & 0.380 & 0.700 & 0.330 & 0.560 & 0.492 \\
Qwen3.5-enc. & 0.340 & 0.630 & 0.480 & 0.770 & 0.555 \\
GLM-4.6V-enc. & 0.430 & 0.750 & 0.410 & 0.730 & 0.580 \\
Gemma 3-enc. & 0.510 & 0.730 & 0.410 & 0.740 & 0.598 \\
\midrule
\multicolumn{6}{l}{\textit{LLMs}} \\
Gemini 3 Flash & 0.590 & 0.790 & 0.550 & 0.890 & 0.705 \\
Gemini 3.1 Flash Lite & 0.600 & \textbf{0.820} & 0.600 & \textbf{0.900} & \textbf{0.730} \\
Qwen3.5 & 0.593 & 0.700 & 0.581 & 0.840 & 0.681 \\
GLM-4.6V & 0.479 & \textbf{0.820} & 0.414 & 0.720 & 0.611 \\
Gemma 3 & 0.480 & 0.600 & 0.420 & 0.720 & 0.555 \\
GPT-5 Nano & 0.550 & 0.680 & 0.410 & 0.720 & 0.590 \\
\bottomrule
\end{tabular}%
}
\end{table}

\begin{table}[!htb]
\centering
\caption{Cross-institution tile relative similarity accuracy on TCGA.}
\label{tab:cross_inst_tcga}
\resizebox{\textwidth}{!}{%
\begin{tabular}{lccccccc}
\toprule
Method & TSS-22 normal & TSS-22 tumor & TSS-56 normal & TSS-56 tumor & TSS-66 normal & TSS-66 tumor & All \\
\midrule
\multicolumn{8}{l}{\textit{Pathology foundation models}} \\
UNI2H & 0.190 & 0.540 & 0.220 & 0.450 & 0.090 & 0.380 & 0.312 \\
H-optimus-0 & 0.020 & 0.040 & 0.200 & 0.300 & 0.200 & 0.300 & 0.177 \\
Midnight & 0.110 & 0.080 & \textbf{0.470} & 0.590 & 0.300 & 0.590 & 0.357 \\
CONCH v1.5 & 0.300 & 0.410 & 0.300 & 0.480 & 0.300 & 0.400 & 0.365 \\
Virchow2 & 0.330 & 0.470 & 0.300 & 0.530 & 0.430 & 0.580 & 0.440 \\
\midrule
\multicolumn{8}{l}{\textit{General foundation models}} \\
Gemini Emb.\ 2 & 0.490 & 0.630 & 0.360 & 0.580 & 0.390 & 0.570 & 0.503 \\
DINOv2 & 0.350 & 0.490 & 0.300 & 0.590 & 0.410 & 0.520 & 0.443 \\
ResNet-50 & 0.450 & 0.510 & 0.250 & 0.390 & 0.370 & 0.640 & 0.435 \\
Qwen3.5-enc. & 0.330 & 0.410 & 0.350 & 0.490 & 0.330 & 0.550 & 0.410 \\
GLM-4.6V-enc. & 0.350 & 0.340 & 0.460 & 0.500 & 0.330 & 0.440 & 0.403 \\
Gemma 3-enc. & 0.140 & 0.390 & 0.380 & 0.600 & 0.260 & 0.540 & 0.385 \\
\midrule
\multicolumn{8}{l}{\textit{LLMs}} \\
Gemini 3 Flash & 0.490 & 0.750 & 0.360 & 0.610 & 0.390 & \textbf{0.680} & 0.547 \\
Gemini 3.1 Flash Lite & \textbf{0.680} & 0.790 & 0.400 & 0.560 & 0.360 & 0.560 & 0.558 \\
Qwen3.5 & 0.609 & \textbf{0.794} & 0.455 & \textbf{0.630} & 0.424 & 0.626 & \textbf{0.589} \\
GLM-4.6V & 0.413 & 0.544 & 0.299 & 0.540 & 0.378 & 0.660 & 0.473 \\
Gemma 3 & 0.490 & 0.640 & 0.430 & 0.580 & \textbf{0.500} & 0.520 & 0.527 \\
GPT-5 Nano & 0.580 & 0.540 & 0.360 & 0.590 & 0.370 & 0.650 & 0.515 \\
\bottomrule
\end{tabular}%
}
\end{table}

\begin{table}[!htb]
\centering
\caption{Cross-institution tile relative similarity accuracy on BEETLE.}
\label{tab:cross_inst_beetle}
\resizebox{\textwidth}{!}{%
\begin{tabular}{lccccccccccccc}
\toprule
Method & NKI inv. & NKI nec. & NKI non-inv. & RUMC inv. & RUMC nec. & RUMC non-inv. & SCH inv. & SCH nec. & SCH non-inv. & TCGA inv. & TCGA nec. & TCGA non-inv. & All \\
\midrule
\multicolumn{14}{l}{\textit{Pathology foundation models}} \\
UNI2H & 0.650 & 0.750 & 0.800 & 0.620 & 0.670 & 0.670 & 0.300 & 0.310 & 0.320 & 0.400 & \textbf{0.680} & 0.520 & 0.557 \\
H-optimus-0 & 0.600 & 0.570 & 0.430 & \textbf{0.800} & 0.410 & 0.490 & 0.410 & 0.310 & 0.390 & 0.500 & 0.170 & 0.280 & 0.447 \\
Midnight & 0.350 & 0.740 & \textbf{0.860} & 0.120 & 0.630 & 0.700 & 0.140 & 0.420 & 0.590 & 0.190 & 0.290 & 0.700 & 0.477 \\
CONCH v1.5 & 0.450 & 0.470 & 0.450 & 0.700 & 0.710 & 0.590 & 0.390 & 0.370 & 0.360 & 0.360 & 0.310 & 0.430 & 0.466 \\
Virchow2 & 0.390 & 0.810 & 0.740 & 0.660 & 0.520 & \textbf{0.850} & 0.600 & 0.170 & 0.480 & 0.420 & 0.420 & \textbf{0.720} & 0.565 \\
\midrule
\multicolumn{14}{l}{\textit{General foundation models}} \\
Gemini Emb.\ 2 & 0.640 & 0.670 & 0.570 & 0.690 & 0.710 & 0.520 & 0.460 & 0.630 & 0.400 & 0.690 & 0.360 & 0.420 & 0.563 \\
DINOv2 & 0.710 & 0.500 & 0.600 & 0.510 & 0.190 & 0.350 & \textbf{0.620} & 0.480 & 0.550 & \textbf{0.800} & 0.370 & 0.440 & 0.510 \\
ResNet-50 & 0.480 & 0.360 & 0.360 & 0.370 & 0.450 & 0.470 & 0.490 & 0.470 & 0.470 & 0.650 & 0.490 & 0.490 & 0.463 \\
Qwen3.5-enc. & 0.590 & 0.230 & 0.340 & 0.460 & 0.160 & 0.260 & 0.290 & 0.430 & 0.460 & 0.400 & 0.170 & 0.160 & 0.329 \\
GLM-4.6V-enc. & 0.550 & 0.400 & 0.670 & 0.600 & 0.380 & 0.510 & 0.460 & 0.350 & 0.410 & 0.660 & 0.350 & 0.430 & 0.481 \\
Gemma 3-enc. & 0.650 & 0.530 & 0.470 & 0.630 & 0.750 & 0.520 & 0.540 & 0.480 & 0.410 & 0.730 & 0.320 & 0.530 & 0.547 \\
\midrule
\multicolumn{14}{l}{\textit{LLMs}} \\
Gemini 3 Flash & \textbf{0.760} & 0.810 & 0.540 & 0.790 & 0.660 & 0.520 & 0.430 & 0.460 & 0.360 & 0.680 & 0.160 & 0.290 & 0.538 \\
Gemini 3.1 Flash Lite & 0.650 & 0.810 & 0.610 & 0.760 & 0.660 & 0.450 & 0.510 & 0.650 & 0.450 & 0.730 & 0.310 & 0.360 & \textbf{0.579} \\
Qwen3.5 & 0.620 & \textbf{0.820} & 0.590 & 0.620 & \textbf{0.770} & 0.590 & 0.460 & \textbf{0.660} & 0.360 & 0.760 & 0.200 & 0.400 & 0.571 \\
GLM-4.6V & 0.707 & 0.500 & 0.600 & 0.660 & 0.354 & 0.379 & 0.550 & 0.530 & 0.560 & 0.720 & 0.020 & 0.378 & 0.497 \\
Gemma 3 & 0.600 & 0.670 & 0.510 & 0.640 & 0.650 & 0.490 & 0.610 & 0.630 & 0.600 & 0.660 & 0.200 & 0.400 & 0.555 \\
GPT-5 Nano & 0.710 & 0.510 & 0.720 & 0.550 & 0.590 & 0.540 & \textbf{0.620} & 0.560 & \textbf{0.700} & 0.590 & 0.300 & 0.450 & 0.570 \\
\bottomrule
\end{tabular}%
}
\end{table}

\FloatBarrier
\subsection{Within-Institution Accuracy}
\label{sec:within_inst_supplementary}

\subsubsection{CAMELYON16}

\begin{table}[!htb]
\centering
\caption{Within-institution tile relative similarity accuracy on CAMELYON16.}
\label{tab:within_inst_camelyon16}
\resizebox{\textwidth}{!}{%
\begin{tabular}{lccccc}
\toprule
Method & Radboud normal & Radboud tumor & Utrecht normal & Utrecht tumor & All \\
\midrule
\multicolumn{6}{l}{\textit{Pathology foundation models}} \\
UNI2H & 0.610 & \textbf{0.970} & 0.530 & 0.940 & 0.763 \\
H-optimus-0 & 0.570 & 0.890 & 0.680 & 0.890 & 0.758 \\
Midnight & \textbf{0.730} & \textbf{0.970} & 0.680 & 0.930 & 0.828 \\
CONCH v1.5 & 0.610 & 0.950 & 0.560 & 0.910 & 0.758 \\
Virchow2 & 0.710 & 0.960 & \textbf{0.790} & \textbf{0.950} & \textbf{0.853} \\
\midrule
\multicolumn{6}{l}{\textit{General foundation models}} \\
DINOv2 & 0.450 & 0.840 & 0.370 & 0.870 & 0.633 \\
ResNet-50 & 0.460 & 0.820 & 0.460 & 0.760 & 0.625 \\
Qwen3.5-enc. & 0.390 & 0.910 & 0.610 & 0.790 & 0.675 \\
GLM-4.6V-enc. & 0.240 & 0.420 & 0.330 & 0.470 & 0.365 \\
Gemma 3-enc. & 0.600 & 0.860 & 0.510 & 0.880 & 0.713 \\
\midrule
\multicolumn{6}{l}{\textit{LLMs}} \\
Gemini 3 Flash & 0.690 & 0.920 & 0.650 & 0.919 & 0.794 \\
Gemini 3.1 Flash Lite & \textbf{0.730} & 0.930 & 0.690 & 0.910 & 0.815 \\
Qwen3.5 & 0.705 & 0.908 & 0.674 & 0.902 & 0.797 \\
GLM-4.6V & 0.625 & 0.907 & 0.505 & 0.870 & 0.732 \\
Gemma 3 & 0.596 & 0.727 & 0.590 & 0.860 & 0.693 \\
GPT-5 Nano & 0.620 & 0.780 & 0.580 & 0.770 & 0.688 \\
\bottomrule
\end{tabular}%
}
\end{table}

\subsubsection{TCGA}

\begin{table}[!htb]
\centering
\caption{Within-institution tile relative similarity accuracy on TCGA.}
\label{tab:within_inst_tcga}
\resizebox{\textwidth}{!}{%
\begin{tabular}{lccccccc}
\toprule
Method & TSS-22 normal & TSS-22 tumor & TSS-56 normal & TSS-56 tumor & TSS-66 normal & TSS-66 tumor & All \\
\midrule
\multicolumn{8}{l}{\textit{Pathology foundation models}} \\
UNI2H & 0.730 & 0.760 & 0.430 & 0.690 & 0.490 & 0.830 & 0.655 \\
H-optimus-0 & 0.720 & 0.820 & 0.450 & 0.730 & 0.460 & \textbf{0.920} & 0.683 \\
Midnight & 0.760 & 0.750 & 0.520 & 0.650 & \textbf{0.570} & 0.890 & 0.690 \\
CONCH v1.5 & 0.710 & 0.600 & 0.450 & 0.690 & 0.470 & 0.870 & 0.632 \\
Virchow2 & 0.780 & 0.720 & 0.500 & 0.720 & 0.540 & 0.840 & 0.683 \\
\midrule
\multicolumn{8}{l}{\textit{General foundation models}} \\
DINOv2 & 0.610 & 0.790 & 0.430 & 0.710 & 0.470 & 0.780 & 0.632 \\
ResNet-50 & 0.660 & 0.700 & 0.410 & 0.650 & 0.440 & 0.790 & 0.608 \\
Qwen3.5-enc. & 0.660 & 0.810 & 0.500 & 0.640 & 0.500 & 0.680 & 0.632 \\
GLM-4.6V-enc. & 0.200 & 0.410 & 0.320 & 0.460 & 0.160 & 0.340 & 0.315 \\
Gemma 3-enc. & 0.760 & 0.800 & 0.470 & 0.660 & 0.560 & 0.680 & 0.655 \\
\midrule
\multicolumn{8}{l}{\textit{LLMs}} \\
Gemini 3 Flash & 0.810 & 0.810 & 0.460 & \textbf{0.840} & 0.510 & 0.870 & \textbf{0.717} \\
Gemini 3.1 Flash Lite & \textbf{0.830} & 0.760 & 0.490 & 0.800 & 0.550 & 0.750 & 0.697 \\
Qwen3.5 & 0.773 & 0.800 & \textbf{0.535} & 0.723 & \textbf{0.570} & 0.816 & 0.702 \\
GLM-4.6V & 0.747 & \textbf{0.828} & 0.429 & 0.710 & 0.458 & 0.810 & 0.666 \\
Gemma 3 & 0.700 & 0.790 & 0.510 & 0.650 & 0.540 & 0.710 & 0.650 \\
GPT-5 Nano & 0.700 & 0.670 & 0.460 & 0.550 & 0.520 & 0.670 & 0.595 \\
\bottomrule
\end{tabular}%
}
\end{table}

\subsubsection{BEETLE}

\begin{table}[!htb]
\centering
\caption{Within-institution tile relative similarity accuracy on BEETLE.}
\label{tab:within_inst_beetle}
\resizebox{\textwidth}{!}{%
\begin{tabular}{lccccccccccccc}
\toprule
Method & NKI inv. & NKI nec. & NKI non-inv. & RUMC inv. & RUMC nec. & RUMC non-inv. & SCH inv. & SCH nec. & SCH non-inv. & TCGA inv. & TCGA nec. & TCGA non-inv. & All \\
\midrule
\multicolumn{14}{l}{\textit{Pathology foundation models}} \\
UNI2H & 0.930 & 0.940 & \textbf{0.980} & 0.770 & \textbf{1.000} & 0.945 & 0.945 & 0.905 & 0.740 & 0.785 & \textbf{1.000} & 0.780 & 0.893 \\
H-optimus-0 & 0.885 & 0.950 & 0.965 & 0.845 & 1.000 & 0.910 & \textbf{0.985} & 0.795 & 0.715 & 0.815 & 1.000 & \textbf{0.795} & 0.888 \\
Midnight & 0.940 & \textbf{0.965} & 0.950 & 0.870 & 1.000 & 0.945 & 0.930 & \textbf{0.905} & 0.770 & 0.905 & 0.995 & 0.755 & \textbf{0.910} \\
CONCH v1.5 & 0.840 & 0.960 & 0.805 & 0.870 & 0.995 & 0.905 & 0.970 & 0.825 & 0.640 & 0.765 & 0.975 & 0.755 & 0.859 \\
Virchow2 & \textbf{0.945} & 0.955 & 0.970 & 0.835 & 1.000 & \textbf{0.975} & 0.980 & 0.870 & \textbf{0.800} & 0.790 & 0.975 & 0.765 & 0.905 \\
\midrule
\multicolumn{14}{l}{\textit{General foundation models}} \\
Gemini Emb.\ 2 & 0.835 & 0.965 & 0.765 & 0.815 & 0.980 & 0.800 & 0.890 & 0.830 & 0.595 & 0.910 & 1.000 & 0.705 & 0.841 \\
DINOv2 & 0.835 & 0.930 & 0.780 & 0.675 & 0.955 & 0.760 & 0.900 & 0.625 & 0.575 & 0.915 & 0.800 & 0.690 & 0.787 \\
ResNet-50 & 0.755 & 0.950 & 0.735 & 0.720 & 0.990 & 0.725 & 0.835 & 0.705 & 0.570 & 0.805 & 0.975 & 0.670 & 0.786 \\
Qwen3.5-enc. & 0.805 & 0.950 & 0.785 & 0.720 & 0.885 & 0.715 & 0.815 & 0.745 & 0.555 & 0.870 & 0.985 & 0.710 & 0.795 \\
GLM-4.6V-enc. & 0.675 & 0.885 & 0.820 & 0.630 & 0.945 & 0.760 & 0.840 & 0.650 & 0.575 & 0.745 & 0.825 & 0.710 & 0.755 \\
Gemma 3-enc. & 0.775 & 0.955 & 0.795 & 0.895 & 0.975 & 0.800 & 0.850 & 0.695 & 0.600 & 0.930 & 0.935 & 0.755 & 0.830 \\
\midrule
\multicolumn{14}{l}{\textit{LLMs}} \\
Gemini 3 Flash & 0.900 & 0.955 & 0.855 & 0.899 & 0.985 & 0.780 & 0.870 & 0.730 & 0.600 & 0.895 & 0.990 & 0.775 & 0.853 \\
Gemini 3.1 Flash Lite & 0.900 & 0.965 & 0.810 & 0.835 & 1.000 & 0.745 & 0.895 & 0.765 & 0.610 & 0.820 & 0.970 & 0.745 & 0.838 \\
Qwen3.5 & 0.838 & 0.964 & 0.820 & \textbf{0.919} & 1.000 & 0.781 & 0.899 & 0.893 & 0.648 & 0.848 & 0.985 & 0.758 & 0.863 \\
GLM-4.6V & 0.798 & 0.943 & 0.868 & 0.850 & 1.000 & 0.795 & 0.895 & 0.719 & 0.582 & \textbf{0.934} & 0.980 & 0.749 & 0.843 \\
Gemma 3 & 0.716 & 0.934 & 0.807 & 0.778 & 0.969 & 0.769 & 0.850 & 0.735 & 0.565 & 0.873 & 0.880 & 0.665 & 0.794 \\
GPT-5 Nano & 0.780 & 0.850 & 0.845 & 0.810 & 0.925 & 0.775 & 0.740 & 0.545 & 0.565 & 0.725 & 0.630 & 0.780 & 0.748 \\
\bottomrule
\end{tabular}%
}
\end{table}

\FloatBarrier
\subsection{Per-Class Pair Confusion}
\label{sec:cross_inst_perclass_appendix}

\begin{figure}[!htb]
\centering
\includegraphics[width=\textwidth]{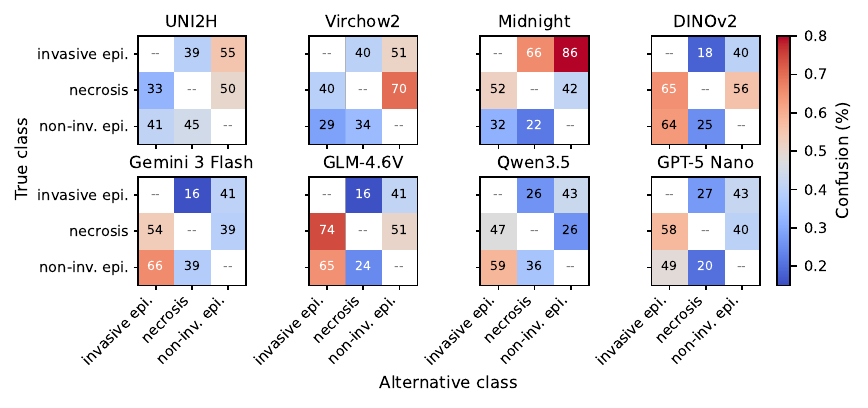}
\caption{Per-class pair confusion for cross-institution comparisons on BEETLE. Rows indicate the positive class $c_+$ and columns the negative class $c_-$. Higher values (warmer colors) indicate greater confusion.}
\label{fig:cross_inst_perclass_beetle}
\end{figure}

\FloatBarrier
\section{Model Cost and Training Data}
\label{sec:cost_training_appendix}

\subsection{Accuracy vs Model Cost}
\label{sec:cost_accuracy_appendix}

\begin{figure}[!htb]
\centering
\includegraphics[width=\textwidth]{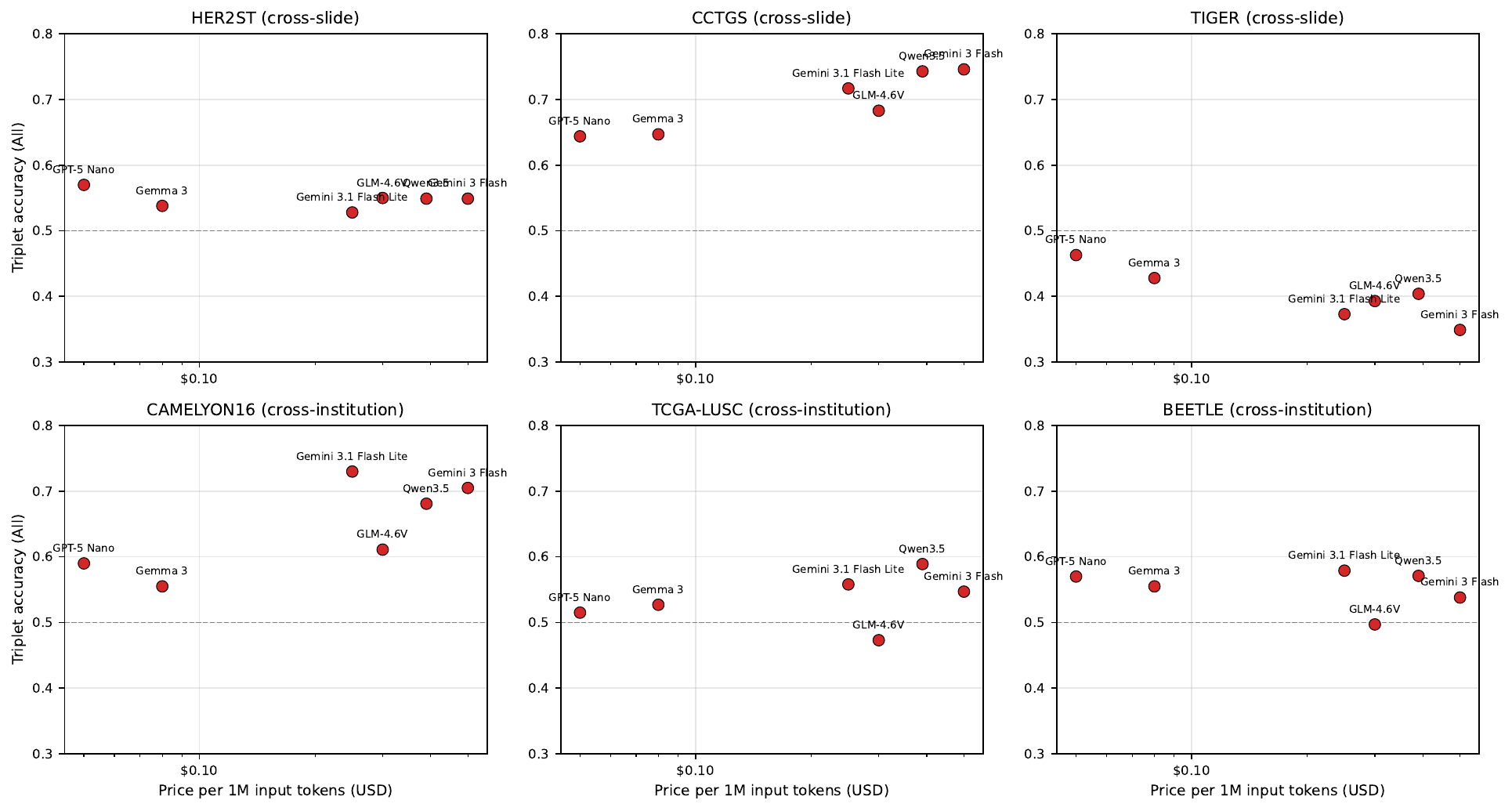}
\caption{LLM relative similarity accuracy vs input-token price. Top row: cross-slide accuracy on HER2ST, CCTGS, and TIGER (Table~\ref{tab:cross_slide_merged}). Bottom row: cross-institution accuracy on CAMELYON16, TCGA, and BEETLE (Tables~\ref{tab:cross_inst_camelyon16}, \ref{tab:cross_inst_tcga}, and~\ref{tab:cross_inst_beetle}). Each point is one LLM. Dashed line indicates chance (0.5).}
\label{fig:cost_accuracy_llm_merged}
\end{figure}

\begin{figure}[!htb]
\centering
\includegraphics[width=\textwidth]{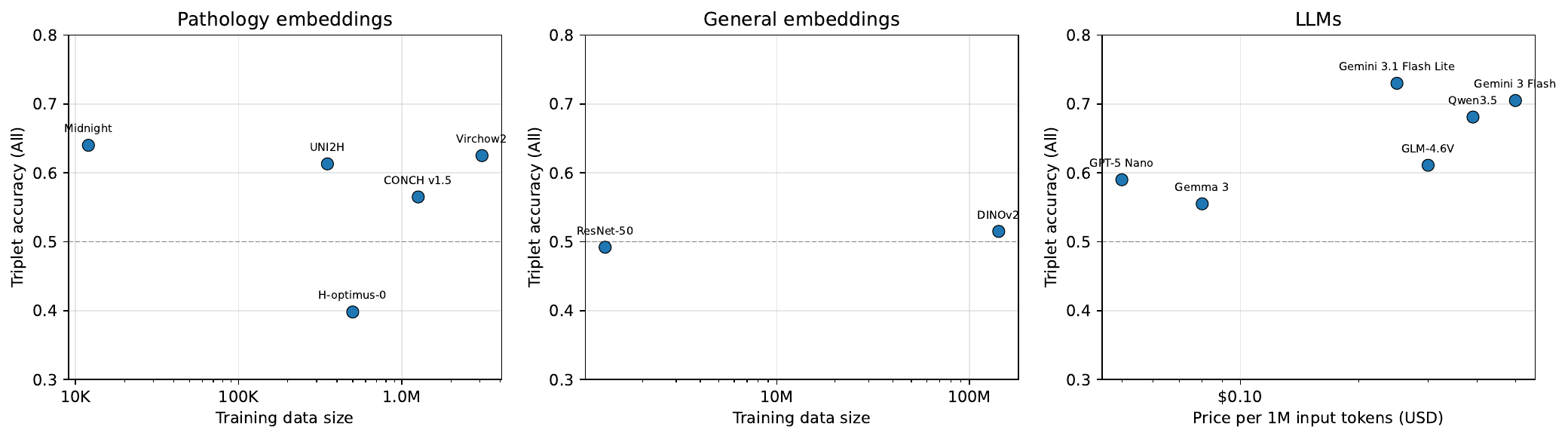}
\caption{CAMELYON16 cross-institution relative similarity accuracy vs model cost. Left: pathology foundation model accuracy vs training data size. Center: general foundation model accuracy vs training data size (Gemini Emb.\ 2 training size estimated at $\sim$1B). Right: LLM accuracy vs price per million input tokens. Dashed line indicates chance (50\%).}
\label{fig:cost_accuracy_camelyon16}
\end{figure}

\begin{figure}[!htb]
\centering
\includegraphics[width=\textwidth]{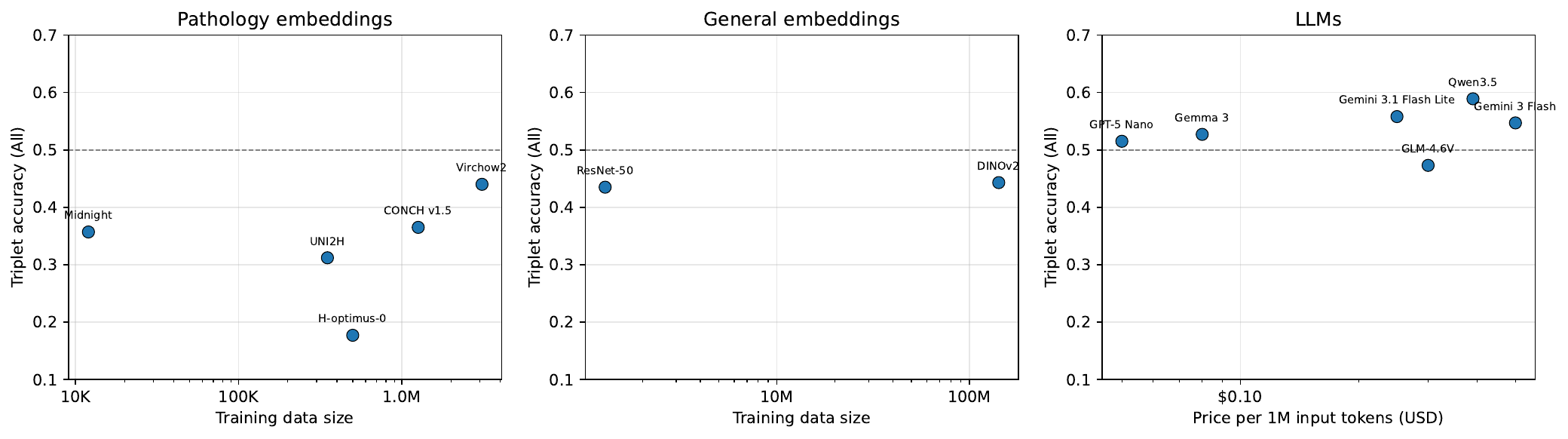}
\caption{TCGA cross-institution relative similarity accuracy vs model cost. Left: pathology foundation model accuracy vs training data size. Center: general foundation model accuracy vs training data size. Right: LLM accuracy vs price per million input tokens. Dashed line indicates chance (50\%).}
\label{fig:cost_accuracy_tcga_lusc}
\end{figure}

\begin{figure}[!htb]
\centering
\includegraphics[width=\textwidth]{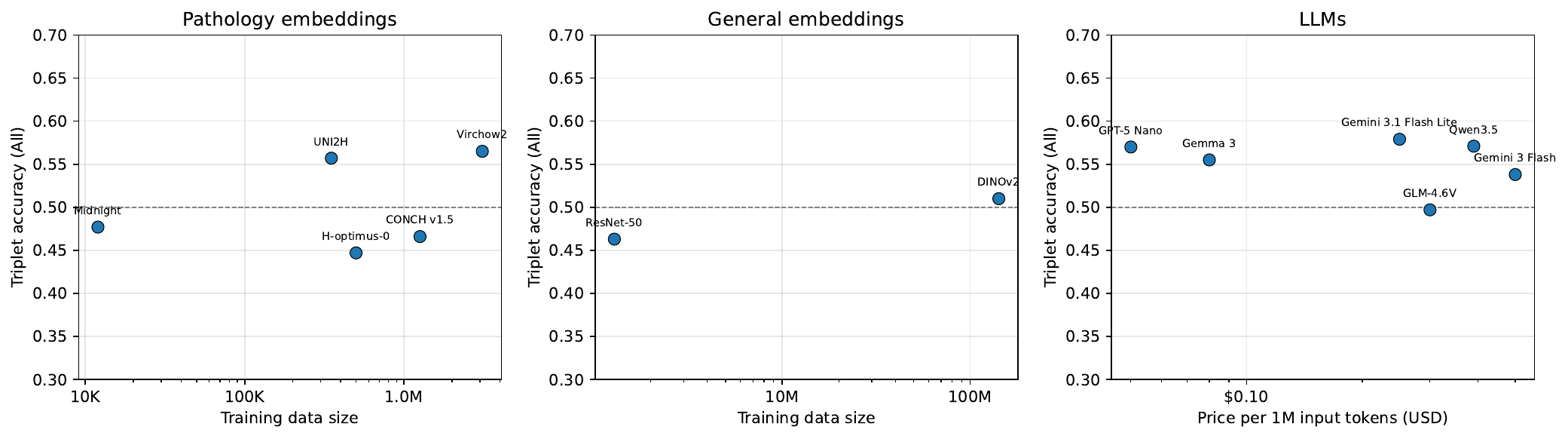}
\caption{BEETLE cross-institution relative similarity accuracy vs model cost. Left: pathology foundation model accuracy vs training data size. Center: general foundation model accuracy vs training data size. Right: LLM accuracy vs price per million input tokens. Dashed line indicates chance (50\%).}
\label{fig:cost_accuracy_beetle}
\end{figure}

\begin{figure}[!htb]
\centering
\includegraphics[width=0.6\textwidth]{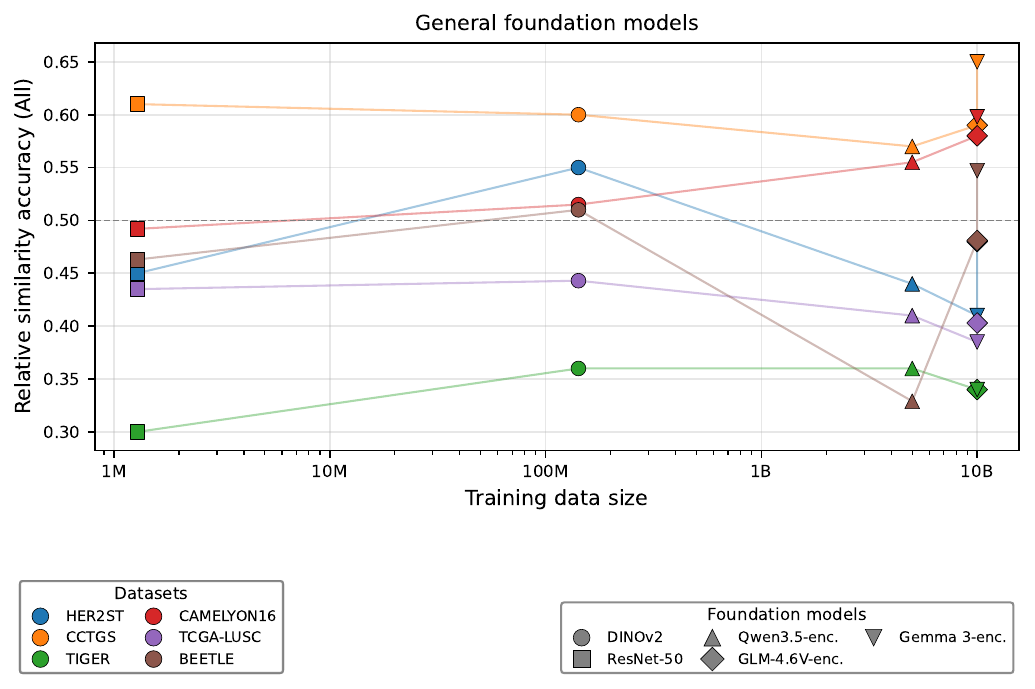}
\caption{General foundation model relative similarity accuracy vs training data size across all six datasets. Color indicates dataset; marker shape indicates method. Lines connect methods within the same dataset. Dashed line indicates chance (0.5). Training data sizes for LLM-derived encoders are estimates based on the upstream vision encoder's pretraining data (Table~\ref{tab:training_data_sources}).}
\label{fig:cost_accuracy_gen_emb}
\end{figure}

\FloatBarrier
\subsection{Training Data Sources}
\label{sec:training_data_sources}

\begin{table}[!htb]
\centering
\caption{Training data sizes for foundation models used in cost/size vs accuracy analysis (Figure~\ref{fig:cost_accuracy_llm_single}). Sizes marked with $\dagger$ are estimates based on the upstream vision encoder's pretraining data, as the multimodal training data size is not separately disclosed.}
\label{tab:training_data_sources}
\resizebox{\textwidth}{!}{%
\begin{tabular}{llllp{7cm}}
\toprule
Model & Training size & Source & URL & Quote \\
\midrule
\multicolumn{5}{l}{\textit{Pathology foundation models}} \\
UNI2H & $>$350K WSIs & HuggingFace model card & \url{https://huggingface.co/MahmoodLab/UNI2-h} & ``Over 200 million image tiles sampled from over 350k diverse H\&E and IHC slides sourced from Mass General Brigham.'' \\
H-optimus-0 & $>$500K WSIs & HuggingFace model card & \url{https://huggingface.co/bioptimus/H-optimus-0} & ``The model is a 1.1B parameter vision transformer trained on a proprietary collection of more than 500,000 H\&E stained whole slide histology images.'' \\
Midnight & 12K WSIs & Tolkach et al.\ (2025) & \url{https://arxiv.org/abs/2504.05186} & ``We trained our first FM on the 12k TCGA WSIs alone.'' \\
CONCH v1.5 & 1.26M pairs & Ding et al.\ (2024) & \url{https://arxiv.org/abs/2411.19666} & ``CONCHv1.5, an extended version of CONCH, which was trained with 1.26 million image-caption pairs using the CoCa training objective.'' \\
Virchow2 & 3.1M WSIs & Zimmermann et al.\ (2024) & \url{https://arxiv.org/abs/2408.00738} & ``each trained with 3.1 million histopathology whole slide images'' \\
\midrule
\multicolumn{5}{l}{\textit{General foundation models}} \\
Gemini Emb.\ 2$^\dagger$ & $\sim$1B images & Lee et al.\ (2025) & \url{https://arxiv.org/abs/2503.07891} & Training data size not disclosed; estimated at $\sim$1B based on the Gemini model family. \\
DINOv2 & 142M images & Oquab et al.\ (2024) & \url{https://arxiv.org/abs/2304.07193} & ``We gathered a small but diverse corpus of 142M images to validate our approach.'' \\
ResNet-50 & 1.28M images & ILSVRC 2012 & \url{https://www.image-net.org/challenges/LSVRC/2012/} & ``The training data, the subset of ImageNet containing the 1000 categories and 1.2 million images'' \\
Qwen3.5-enc.$^\dagger$ & $\sim$5B images & DFN-5B model card & \url{https://huggingface.co/apple/DFN5B-CLIP-ViT-H-14-378} & ``trained on a 5B subset of a pool of 43B uncurated image-text pairs'' \\
GLM-4.6V-enc.$^\dagger$ & $\sim$10B pairs & Yang et al.\ (2025) & \url{https://arxiv.org/abs/2507.01006} & ``an initial pool of over 10 billion image-text pairs from diverse sources'' \\
Gemma 3-enc.$^\dagger$ & $\sim$10B pairs & SigLIP model card; Gemma 3 report & \url{https://arxiv.org/abs/2503.19786} & ``a 400M variant of the SigLIP encoder''; SigLIP pre-trained on WebLI ($\sim$10B image-text pairs). \\
\bottomrule
\end{tabular}%
}
\end{table}

\FloatBarrier
\subsection{LLM Input-Token Prices}
\label{sec:llm_prices}

\begin{table}[!htb]
\centering
\caption{LLM input-token prices used in cost vs accuracy analysis (Figure~\ref{fig:cost_accuracy_llm_single}).}
\label{tab:llm_prices}
\begin{tabular}{lc}
\toprule
Model & Price per 1M input tokens (USD) \\
\midrule
Gemini 3 Flash & \$0.50 \\
Gemini 3.1 Flash Lite & \$0.25 \\
Qwen3.5 & \$0.39 \\
GLM-4.6V & \$0.30 \\
Gemma 3 & \$0.08 \\
GPT-5 Nano & \$0.05 \\
\bottomrule
\end{tabular}
\end{table}

\clearpage
\section{Ablation Studies}
\label{sec:ablations_appendix}

\subsection{Prompt Ablation}
\label{sec:prompt_ablation_appendix}

We compare two prompts for LLM-based relative similarity evaluation. In both cases, the LLM receives three images (one reference and two candidates) and must select which candidate is more similar to the reference.

\paragraph{Biology-focused prompt (default).}
\begin{quote}
\small
Attached are three H\&E histology image tiles: one reference tile (reference\_tile) and two candidate tiles (candidate\_tile\_1, candidate\_tile\_2).
Your task is to determine which candidate tile is more biologically similar to the reference tile.
When assessing similarity, focus on biological differences:
\begin{itemize}[nosep]
\item Cell morphology (e.g., cell size, shape, nuclear features)
\item Tissue architecture (e.g., glandular organization, stromal patterns, cell density)
\end{itemize}
Ignore non-biological differences such as imaging artifacts, stain intensity, resolution, or tissue preparation variations.
\end{quote}

\paragraph{Minimal prompt (ablation).}
\begin{quote}
\small
Attached are three images: one reference image (reference\_tile) and two candidate images (candidate\_tile\_1, candidate\_tile\_2).
Which candidate is more similar to the reference?
\end{quote}

Both prompts require the LLM to output a JSON response with a \texttt{more\_similar\_candidate} field. The biology-focused prompt provides domain-specific guidance on which features to attend to and which to ignore, while the minimal prompt contains no such guidance.

\clearpage
\section{Statistical Significance}
\label{sec:statistical_significance}

We report bootstrap 95\% confidence intervals and McNemar's test results for HER2ST cross-slide (2,000 comparisons) and CAMELYON16 cross-institution (400 comparisons), the two experiments where all models evaluate the same comparison tuples. For LLMs that fail to return a valid response on some comparisons, McNemar's test uses only the subset where both models returned valid responses (the $n$ column in Table~\ref{tab:statistical_significance} reports valid counts per model). Bootstrap confidence intervals are computed with 10,000 resamples over the sampled comparisons. For McNemar's test, we use the exact binomial test when the number of discordant pairs is below 25, and the chi-squared test with continuity correction otherwise. The pairwise comparisons in Table~\ref{tab:mcnemar_tests} are exploratory (best model selected post hoc within each category), so p-values should be interpreted accordingly.

\begin{table}[!htb]
\centering
\caption{Bootstrap 95\% confidence intervals for overall accuracy on HER2ST (cross-slide, $n = 2000$) and CAMELYON16 (cross-institution, $n = 400$). Models are grouped by category.}
\label{tab:statistical_significance}
\resizebox{\textwidth}{!}{%
\begin{tabular}{l ccc ccc}
\toprule
& \multicolumn{3}{c}{HER2ST (cross-slide)} & \multicolumn{3}{c}{CAMELYON16 (cross-inst.)} \\
\cmidrule(lr){2-4} \cmidrule(lr){5-7}
Method & Acc. & 95\% CI & $n$ & Acc. & 95\% CI & $n$ \\
\midrule
\multicolumn{7}{l}{\textit{Pathology foundation models}} \\
UNI2H & 0.459 & [0.437, 0.480] & 2000 & 0.613 & [0.565, 0.660] & 400 \\
H-optimus-0 & 0.439 & [0.416, 0.461] & 2000 & 0.398 & [0.350, 0.445] & 400 \\
Midnight & 0.464 & [0.442, 0.485] & 2000 & 0.640 & [0.593, 0.685] & 400 \\
CONCH v1.5 & 0.450 & [0.428, 0.472] & 2000 & 0.565 & [0.517, 0.613] & 400 \\
Virchow2 & 0.459 & [0.436, 0.480] & 2000 & 0.625 & [0.578, 0.672] & 400 \\
\midrule
\multicolumn{7}{l}{\textit{General foundation models}} \\
Gemini Emb.\ 2 & 0.482 & [0.461, 0.504] & 2000 & 0.610 & [0.562, 0.657] & 400 \\
DINOv2 & 0.549 & [0.528, 0.571] & 2000 & 0.515 & [0.465, 0.562] & 400 \\
ResNet-50 & 0.453 & [0.431, 0.474] & 2000 & 0.492 & [0.443, 0.540] & 400 \\
Qwen3.5-enc. & 0.436 & [0.414, 0.458] & 2000 & 0.555 & [0.505, 0.605] & 400 \\
GLM-4.6V-enc. & 0.476 & [0.454, 0.497] & 2000 & 0.580 & [0.530, 0.630] & 400 \\
Gemma 3-enc. & 0.412 & [0.391, 0.434] & 2000 & 0.598 & [0.550, 0.645] & 400 \\
\midrule
\multicolumn{7}{l}{\textit{LLMs}} \\
Gemini 3 Flash & 0.549 & [0.527, 0.572] & 2000 & 0.705 & [0.660, 0.748] & 400 \\
Gemini 3.1 Flash Lite & 0.528 & [0.506, 0.550] & 2000 & 0.730 & [0.685, 0.775] & 400 \\
Qwen3.5 & 0.549 & [0.527, 0.571] & 2000 & 0.681 & [0.633, 0.729] & 354 \\
GLM-4.6V & 0.550 & [0.529, 0.572] & 1995 & 0.611 & [0.559, 0.659] & 370 \\
Gemma 3 & 0.538 & [0.516, 0.559] & 2000 & 0.555 & [0.507, 0.603] & 400 \\
GPT-5 Nano & 0.570 & [0.547, 0.591] & 2000 & 0.590 & [0.540, 0.637] & 400 \\
\bottomrule
\end{tabular}%
}
\end{table}

\begin{table}[!htb]
\centering
\caption{McNemar's test for key pairwise comparisons. Each test uses paired comparisons where both models returned valid responses. The best LLM is compared against the best pathology foundation model and the best general foundation model for each experiment.}
\label{tab:mcnemar_tests}
\footnotesize
\begin{tabular}{llcccc}
\toprule
Experiment & Comparison & Acc.\ A & Acc.\ B & $\chi^2$ & $p$-value \\
\midrule
\multirow{2}{*}{HER2ST (cross-slide)} & GPT-5 Nano vs Midnight & 0.570 & 0.464 & 58.1 & $<$0.0001 \\
 & GPT-5 Nano vs DINOv2 & 0.570 & 0.549 & 2.0 & 0.153 \\
\midrule
\multirow{2}{*}{CAMELYON16 (cross-inst.)} & Gem.\ 3.1 F.\ Lite vs Midnight & 0.730 & 0.640 & 11.8 & $<$0.001 \\
 & Gem.\ 3.1 F.\ Lite vs Gemini Emb.\ 2 & 0.730 & 0.610 & 26.9 & $<$0.0001 \\
\bottomrule
\end{tabular}
\end{table}

\clearpage
\section{Class Abbreviations}
\label{sec:class_abbreviations}

\begin{table}[!htb]
\centering
\caption{Tissue class abbreviations used in Table~\ref{tab:cross_slide_merged}.}
\label{tab:class_abbreviations}
\begin{tabular}{lll}
\toprule
Dataset & Abbreviation & Full name \\
\midrule
\multirow{5}{*}{HER2ST} & Adi & Adipose \\
 & Bre & Breast glandular \\
 & Can & Cancer \\
 & Con & Connective \\
 & Imm & Immune infiltrate \\
\midrule
\multirow{6}{*}{CCTGS} & Adi & Adipose \\
 & Lam & Lamina propria \\
 & Lym & Lymphovascular invasion \\
 & Mus & Muscularis propria \\
 & Nor & Normal mucosa \\
 & Tum & Tumor \\
\midrule
\multirow{4}{*}{TIGER} & Hea & Healthy glands \\
 & Nec & Necrosis \\
 & Str & Stroma \\
 & Tum & Tumor \\
\bottomrule
\end{tabular}
\end{table}

\clearpage
\section{Model Licenses}
\label{sec:model_licenses}

\begin{table}[h]
\centering
\caption{Licenses for models used in this paper.}
\label{tab:model_licenses}
\begin{tabular}{ll}
\toprule
Model & License \\
\midrule
\multicolumn{2}{l}{\textit{Pathology foundation models}} \\
UNI2H & CC-BY-NC-ND 4.0 \\
H-optimus-0 & Apache 2.0 \\
Midnight & MIT \\
CONCH v1.5 & CC-BY-NC-ND 4.0 \\
Virchow2 & CC-BY-NC-ND 4.0 \\
\midrule
\multicolumn{2}{l}{\textit{General foundation models}} \\
DINOv2 & Apache 2.0 \\
ResNet-50 & BSD 3-Clause \\
Gemini Embedding 2 & Google API Terms of Service \\
Qwen3.5-encoder & Apache 2.0 \\
GLM-4.6V-encoder & MIT \\
Gemma 3-encoder & Gemma Terms of Use \\
\midrule
\multicolumn{2}{l}{\textit{LLMs}} \\
Gemini 3 Flash & Google API Terms of Service \\
Gemini 3.1 Flash Lite & Google API Terms of Service \\
GPT-5 Nano & OpenAI Terms of Service \\
Gemma 3 & Gemma Terms of Use \\
GLM-4.6V & MIT \\
Qwen3.5 & Apache 2.0 \\
\bottomrule
\end{tabular}
\end{table}

\newpage
\section*{NeurIPS Paper Checklist}

\begin{enumerate}

\item {\bf Claims}
    \item[] Question: Do the main claims made in the abstract and introduction accurately reflect the paper's contributions and scope?
    \item[] Answer: \answerYes{}
    \item[] Justification: The abstract and introduction (Section~\ref{sec:introduction}) clearly state four contributions: (1) exposing a robustness gap in pathology foundation models, (2) introducing a relative similarity evaluation framework, (3) demonstrating that multimodal LLMs achieve robust cross-domain similarity, and (4) releasing a benchmark suite. All claims are supported by the experimental results in Section~\ref{sec:results}.

\item {\bf Limitations}
    \item[] Question: Does the paper discuss the limitations of the work performed by the authors?
    \item[] Answer: \answerYes{}
    \item[] Justification: The conclusion discusses limitations: the evaluation is restricted to H\&E-stained images, and future work could extend the framework to other pathology imaging modalities (e.g., immunohistochemistry, multiplexed imaging) and spatial biology data.

\item {\bf Theory assumptions and proofs}
    \item[] Question: For each theoretical result, does the paper provide the full set of assumptions and a complete (and correct) proof?
    \item[] Answer: \answerNA{}
    \item[] Justification: The paper does not include theoretical results or proofs. It is an empirical study evaluating model robustness via a relative similarity framework.

    \item {\bf Experimental result reproducibility}
    \item[] Question: Does the paper fully disclose all the information needed to reproduce the main experimental results of the paper to the extent that it affects the main claims and/or conclusions of the paper (regardless of whether the code and data are provided or not)?
    \item[] Answer: \answerYes{}
    \item[] Justification: The paper describes the evaluation framework (Section~\ref{sec:method}), all datasets with preprocessing details (Section~\ref{sec:benchmark}), all evaluated models by name and version, triplet sampling procedures (Appendix~\ref{sec:experiment_setup}), and the LLM prompt text (Appendix~\ref{sec:prompt_ablation_appendix}). The benchmark triplets, evaluation code, and baseline outputs will be released.

\item {\bf Open access to data and code}
    \item[] Question: Does the paper provide open access to the data and code, with sufficient instructions to faithfully reproduce the main experimental results, as described in supplemental material?
    \item[] Answer: \answerYes{}
    \item[] Justification: Code and data will be released upon acceptance, as stated in the abstract. The paper describes the evaluation framework in sufficient detail for independent reimplementation.

\item {\bf Experimental setting/details}
    \item[] Question: Does the paper specify all the training and test details (e.g., data splits, hyperparameters, how they were chosen, type of optimizer) necessary to understand the results?
    \item[] Answer: \answerYes{}
    \item[] Justification: The paper describes all evaluation details: dataset composition and preprocessing (Section~\ref{sec:benchmark}), triplet construction and sampling procedures (Appendix~\ref{sec:experiment_setup}), embedding distance computation (Section~\ref{sec:method}), and LLM prompt design (Appendix~\ref{sec:prompt_ablation_appendix}). No model training is performed.

\item {\bf Experiment statistical significance}
    \item[] Question: Does the paper report error bars suitably and correctly defined or other appropriate information about the statistical significance of the experiments?
    \item[] Answer: \answerYes{}
    \item[] Justification: Bootstrap 95\% confidence intervals and McNemar's tests are reported in Appendix~\ref{sec:statistical_significance} for HER2ST (cross-slide) and CAMELYON16 (cross-institution). Key pairwise comparisons are tested for statistical significance and referenced in the main text.

\item {\bf Experiments compute resources}
    \item[] Question: For each experiment, does the paper provide sufficient information on the computer resources (type of compute workers, memory, time of execution) needed to reproduce the experiments?
    \item[] Answer: \answerYes{}
    \item[] Justification: The paper reports that all embedding extraction and local model inference were performed on an NVIDIA RTX 4090 GPU (Section~\ref{sec:method}). LLM costs are reported via input-token prices (Appendix~\ref{sec:llm_prices}).

\item {\bf Code of ethics}
    \item[] Question: Does the research conducted in the paper conform, in every respect, with the NeurIPS Code of Ethics \url{https://neurips.cc/public/EthicsGuidelines}?
    \item[] Answer: \answerYes{}
    \item[] Justification: The research uses publicly available datasets and pre-trained models for evaluation purposes. No new data collection involving human subjects was conducted. The work aims to improve the safety and reliability of computational pathology models.

\item {\bf Broader impacts}
    \item[] Question: Does the paper discuss both potential positive societal impacts and negative societal impacts of the work performed?
    \item[] Answer: \answerYes{}
    \item[] Justification: The paper discusses the clinical danger of relying on foundation models whose embeddings are entangled with acquisition context (Section~\ref{sec:introduction}), highlighting negative societal impacts of deploying non-robust models in multi-site clinical settings. Positive impacts are discussed in the conclusion, where LLMs are proposed as complementary tools for cross-institutional retrieval, dataset harmonization, and quality control.

\item {\bf Safeguards}
    \item[] Question: Does the paper describe safeguards that have been put in place for responsible release of data or models that have a high risk for misuse (e.g., pre-trained language models, image generators, or scraped datasets)?
    \item[] Answer: \answerNA{}
    \item[] Justification: The released assets (comparison triplets, evaluation code, baseline outputs) are evaluation tools that do not pose risks for misuse. No new models or scraped datasets are released.

\item {\bf Licenses for existing assets}
    \item[] Question: Are the creators or original owners of assets (e.g., code, data, models), used in the paper, properly credited and are the license and terms of use explicitly mentioned and properly respected?
    \item[] Answer: \answerYes{}
    \item[] Justification: All datasets and models are cited with their original publications (Sections~\ref{sec:benchmark} and~\ref{sec:method}). Model licenses are listed in Appendix~\ref{sec:model_licenses} (Table~\ref{tab:model_licenses}).

\item {\bf New assets}
    \item[] Question: Are new assets introduced in the paper well documented and is the documentation provided alongside the assets?
    \item[] Answer: \answerYes{}
    \item[] Justification: The benchmark is documented in the paper: triplet construction procedures (Sections~\ref{sec:method} and~\ref{sec:benchmark}), sampling details (Appendix~\ref{sec:experiment_setup}), and baseline results. Documentation and evaluation code will accompany the released benchmark.

\item {\bf Crowdsourcing and research with human subjects}
    \item[] Question: For crowdsourcing experiments and research with human subjects, does the paper include the full text of instructions given to participants and screenshots, if applicable, as well as details about compensation (if any)?
    \item[] Answer: \answerNA{}
    \item[] Justification: The paper does not involve crowdsourcing or research with human subjects.

\item {\bf Institutional review board (IRB) approvals or equivalent for research with human subjects}
    \item[] Question: Does the paper describe potential risks incurred by study participants, whether such risks were disclosed to the subjects, and whether Institutional Review Board (IRB) approvals (or an equivalent approval/review based on the requirements of your country or institution) were obtained?
    \item[] Answer: \answerNA{}
    \item[] Justification: The paper does not involve research with human subjects. All data used consists of publicly available histopathology image datasets.

\item {\bf Declaration of LLM usage}
    \item[] Question: Does the paper describe the usage of LLMs if it is an important, original, or non-standard component of the core methods in this research? Note that if the LLM is used only for writing, editing, or formatting purposes and does \emph{not} impact the core methodology, scientific rigor, or originality of the research, declaration is not required.
    \item[] Answer: \answerYes{}
    \item[] Justification: LLMs are a core component of the methodology. Section~\ref{sec:method} describes how multimodal LLMs are used as zero-shot judges for relative similarity evaluation, and Section~\ref{sec:results} reports their performance. All LLM models are named with specific versions.

\end{enumerate}

\end{document}